\documentclass{article}

\usepackage[preprint]{neurips_2026}
\usepackage{amsmath}
\usepackage{enumitem}
\usepackage{graphicx}
\usepackage{natbib}
\usepackage{graphicx}
\usepackage{amsmath,amssymb} 
\usepackage{wrapfig}
\usepackage{multirow}
\usepackage{booktabs}
\usepackage[utf8]{inputenc}
\usepackage{subcaption}
\usepackage{amsthm}
\usepackage[dvipsnames]{xcolor}
\usepackage{subcaption}
\usepackage{hyperref}
\usepackage{tikz}
\usetikzlibrary{positioning,arrows.meta,calc}
\hypersetup{
    breaklinks=true,
    colorlinks=true,
    allcolors=DeepPink
}

\usepackage[utf8]{inputenc} 
\usepackage[T1]{fontenc}    
\usepackage{hyperref}       
\usepackage{url}            
\usepackage{booktabs}       
\usepackage{amsfonts}       
\usepackage{nicefrac}       
\usepackage{microtype}      
\usepackage{xcolor}         
\usepackage{amsmath}
\usepackage{graphicx}
\usepackage{multirow}
\usepackage{tabularx}
\usepackage{booktabs}
\usepackage{caption}
\usepackage[table]{xcolor}
\usepackage{pifont}
\usepackage{tcolorbox}
\usepackage{wrapfig} 

\newcommand{\ie}{\textit{i.e.,}  }

\title{\textcolor{DeepPink}{SphMind:} Towards Robust, Training-Free VLM-based Spatial Reasoning with a 360 Camera}

\author{%
  Shriram Damodaran \\
  EmPACT LAb @ NTU Singapore \\
  \texttt{shriram003@e.ntu.edu.sg} \\
  \And
  Soumyaratna Debnath \\
  EmPACT LAb @ NTU Singapore \\
  \texttt{soumyara004@e.ntu.edu.sg} \\
  \And
  Cheston Tan \\
  IHPC, A*STAR, Singapore \\
  \texttt{cheston\_tan@a-star.edu.sg} \\
  \And
  Addison Lin Wang~\thanks{Corresponding Author} \\
  EmPACT LAb @ NTU Singapore \\
  \texttt{linwang@ntu.edu.sg} \\
}

\definecolor{DeepPink}{rgb}{1,0.078,0.576}

\usepackage{hyperref}
\hypersetup{
    breaklinks=true,
    colorlinks=true,
    allcolors=DeepPink
}

\newtcolorbox{rqbox}{
  colback=blue!5,
  colframe=blue!60,
  boxrule=0pt,
  left=8pt, right=6pt, top=6pt, bottom=6pt
}
\newtcolorbox{takeawaybox}{
  colback=orange!8,
  colframe=orange!70!black,
  boxrule=0pt,
  left=8pt, right=6pt, top=6pt, bottom=6pt
}

\makeatletter
\renewcommand{\@toptitlebar}{}
\renewcommand{\@bottomtitlebar}{}
\makeatother

\begin{document}

\begin{tikzpicture}[remember picture, overlay]

\node[anchor=north west, inner sep=0pt]
at ($(current page.north west)+(0.72in,-0.43in)$)
{
    \includegraphics[
        width=3.0cm,
        trim=3 3 3 3,
        clip
    ]{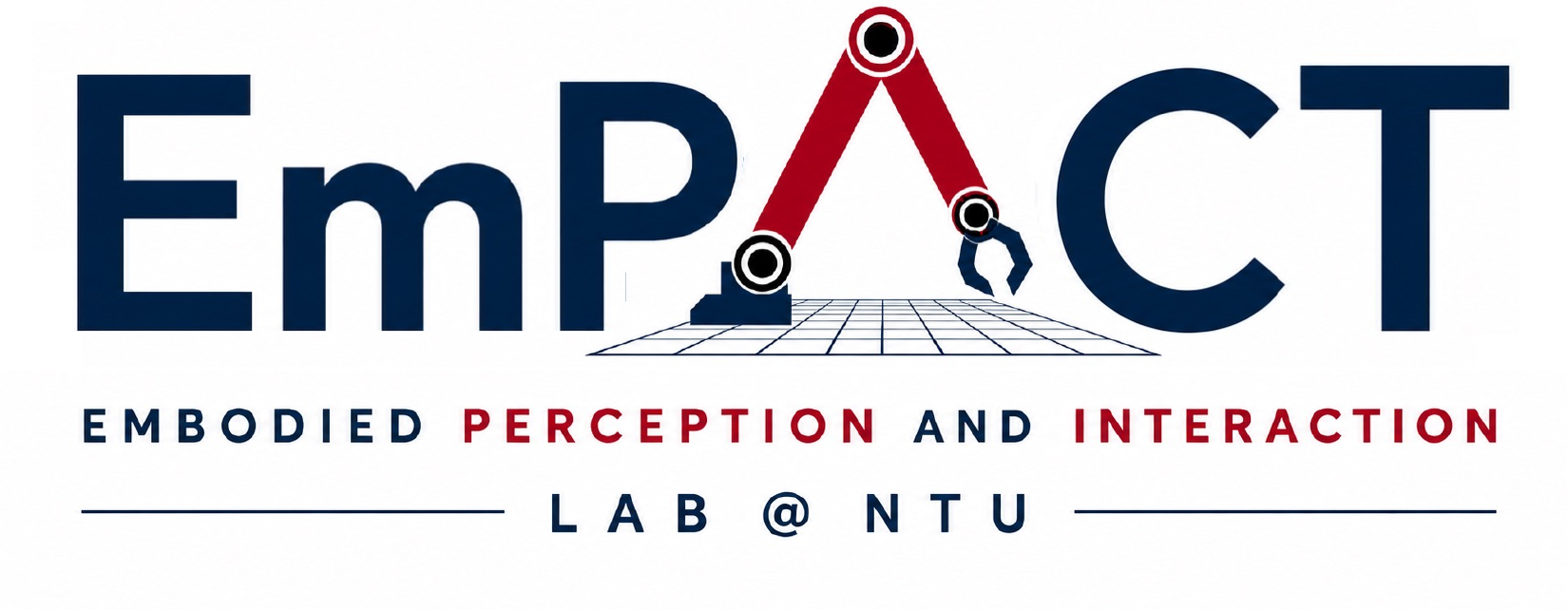}
};

\draw[
    black!20,
    line width=0.35pt
]
($(current page.north west)+(0.72in,-0.98in)$)
--
($(current page.north east)+(-0.72in,-0.98in)$);

\end{tikzpicture}

\maketitle

\begin{figure}[h!]
\label{sec:teaser}
    \centering
    \vspace{-10pt}
    \includegraphics[width=.95\linewidth]{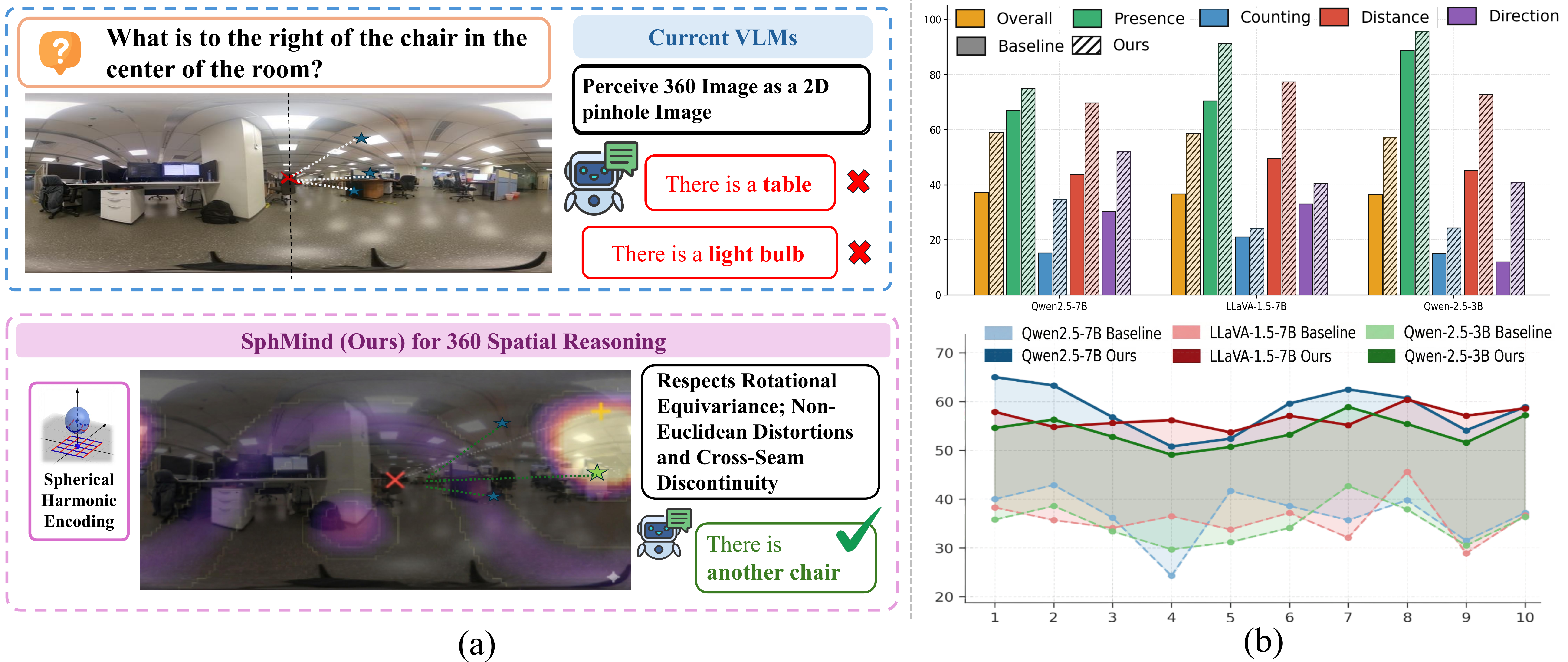}
    \vspace{-3pt}
    \caption{
(\textbf{a}) Given `in-the-wild' real-world 360 capture, existing VLMs fail at spherical reasoning, leading to incorrect spatial predictions. Our SphMind can enable VLMs to generalize to non-Euclidean 3D spaces without retraining.
(\textbf{b}) Accuracy comparison of VLMs vs.\ SphMind across four categories; bars show per-category gains, shaded line shows min--max range across models. \\ Project Page: \textcolor{DeepPink}{\href{https://empactlab.github.io/SphMind-NeurIPS-2026/}{https://empactlab.github.io/SphMind-NeurIPS-2026/}}.
}
    \label{fig:teaser-fig}
\end{figure}

\begin{abstract}
Omnidirectional or 360 cameras provide embodied AI agents, with a holistic, wide field-of-view (FoV) sensing of 3D structures of their surroundings. This advantage has sparked significant interest in leveraging emerging Multi-modal Large Language Models (MLLMs) for omnidirectional spatial reasoning. However, as MLLMs are predominantly trained on 2D perspective images, they struggle with the severe distortions and wrap-around discontinuity inherent to spherical geometry. Consequently, enabling MLLMs to generalize to non-Euclidean 3D spaces without retraining remains a significant open challenge.
In this paper, we propose \textbf{\textsc{SphMind}}, a \textbf{novel}, \textbf{training-free}, and \textbf{plug-and-play} framework that explicitly bridges this gap. \textbf{Our key idea} is to decouple semantic perception from geometric reasoning: rather than forcing MLLMs to internalize complex spherical laws, we preserve their strong semantic priors and handle the geometry externally. To achieve this, we introduce a Spherical Harmonics-based Spatial Graph (\textbf{SHSG}) to map spatial relationships using equivariant transformations. We then couple this with Inference-Time Geometric Grounding (\textbf{IGG}), a \textbf{model-agnostic}, closed-loop optimization process that aligns the MLLM's internal representations with these spherical constraints during inference.
Extensive experiments across three benchmark datasets demonstrate 
the effectiveness of \textsc{SphMind}. Without requiring any 
additional training, our framework achieves over \textbf{21.4\%} 
average gain in directional reasoning on MP3D and Stanford2D-3D, outperforms all 
prompt-engineering strategies by \textbf{+8.7\%} on real-world 
ODI-Bench, and achieves \textbf{$\times$5.9} full rotational 
invariance over baselines under panorama rotation 
all without any dataset-specific tuning.  We further validate on an in-the-wild capture, where \textsc{SphMind} correctly resolves 
directional queries that baseline VLMs fail.
\end{abstract}

\section{Introduction}

Omnidirectional images (ODIs)\footnote{Often represented in Equirectangular Projection (ERP) format.} have gained attention due to their holistic \(180 \times 360\) field-of-view (FOV) \cite{jung2025edmequirectangularprojectionorienteddense,zhang2022deep360survey}.  ODIs provide global spatial awareness beyond the restricted front-facing view of conventional inputs \cite{ropero2026riemindgeometrygroundedspatialagent} important for embodied robotic systems operating in complex 3D environments~\cite{damodaran2026eagor, debnath2026adaptopass}. While 360 cameras provide the context required for 3D scene understanding, interpreting ODIs remains highly challenging. Despite advances in Multi-modal Large Language Models (MLLMs) on perspective 2D images~\cite{li2023evaluatingobjecthallucinationlarge,bai2023qwentechnicalreport,alayrac2022flamingovisuallanguagemodel,debnath2026llmind}, these models suffer severe degradation on ODIs. The core issue is a \textit{spatial-semantic gap}~\cite{wu2025indooropenworldrevealing,10546335}: \textit{MLLMs are essentially ``flat'' world models trained on rectilinear images in \(\mathbb{R}^2\)}. Their architectures rely on 2D Euclidean mechanisms such as translation-invariant convolutions and grid-based attention~\cite{wang2018omnidirectionalcnnvisualplace,cohen2018sphericalcnns,he2025positioneuclideanfoundation}. In contrast, ERP projection introduces non linear distortions and spherical periodicity creates cross-seam boundaries.~\cite{wang2018omnidirectionalcnnvisualplace,cohen2018sphericalcnns,he2025positioneuclideanfoundation}.

Thus, omnidirectional reasoning requires overcoming two coupled challenges: \textbf{1)} modeling continuous non-Euclidean spherical geometry,and \textbf{2)} injecting these constraints into MLLMs without overriding semantic priors. Recent approaches adapt to ODIs via specialized fine-tuning or spherical-to-2D patch decomposition~\cite{chou2020visualquestionanswering360deg,reyarea2022360monodepthhighresolution360degmonocular,Jiang_2021}. However, fine-tuning is data-intensive and generalizes poorly, while patch decomposition breaks spherical topology and weakens global spatial relationships.

To address the challenges, {\textbf{our key idea}} is to \textit{decouple semantic perception from geometric reasoning}. Instead of imposing rigid Euclidean assumptions encoded into MLLM backbones \cite{guo2025flatlandsunlockingspatialintelligence}, we delegate complex spatial transformations and constraints to geometric operations. This frees the MLLM to focus on what it does best: semantic understanding.
while treating geometric grounding as continuous constrained optimization rather than a discrete post-processing heuristic. 
As shown in Fig.~\ref{fig:teaser-fig}(a), current VLMs 
ambiguously assign any object to the right of an anchor as 
``right'', whereas grounding on the sphere via SH yields a 
distinct peak at the correct answer with lower scores 
for distractors.

We propose \textbf{\textsc{SphMind}}, a \textbf{training-free} 
framework with two key components.. \textbf{First}, to understand non-Euclidean geometry, we introduce the Spherical Harmonics-based Spatial Graph (\textbf{SHSG}) to enable reasoning directly on spherical surfaces. Within SHSG, spatial relations are formulated as continuous directional tensor fields rather than discrete pairwise links. By encoding object nodes using spherical harmonics (SH), we preserve directional asymmetry (\ie $A\to B \neq B\to A$) and achieve rotational equivariance \cite{esteves2018learningso3equivariantrepresentations}.  \textbf{Second}, to dynamically bridge the semantic-geometric gap, we develop Inference-Time Geometric Grounding (\textbf{IGG}) (Sec. \ref{sec:igg}). IGG optimizes the MLLM's final-token hidden state to minimize a differentiable geometric 
violation energy, steering the latent representation 
toward geometric consistency while preserving pretrained 
visual priors enabling geometrically-grounded, 
zero-shot decoding without weight updates.

We evaluate \textsc{SphMind} on a 360 spatial 
VQA benchmark of ${\sim}$12K question-answer pairs 
derived analytically from 
Matterport3D~\cite{bata1126} and 
Stanford2D3D~\cite{armeni2017joint2d3dsemanticdataindoor}. 
Across \textbf{three} MLLM backbones, without any  
training (Fig.~\ref{fig:teaser-fig}b), 
\textsc{SphMind} achieves a mean gain of 
$\mathbf{+21.4\%}$ overall, $\mathbf{+20.5\%}$ 
on directional reasoning, and $\mathbf{+18\%}$ 
on distance understanding over the MLLM baselines. 
Under azimuthal rotations of the panorama origin, 
\textsc{SphMind} achieves $\mathbf{{\times}5.9}$ 
full rotational consistency over the ERP-pixel 
baseline, confirming that  
spherical harmonic scoring produces stable predictions 
regardless of spherical orientation. We further 
evaluate zero-shot on 
ODI-Bench~\cite{yang2026odibenchmllmsunderstandimmersive}, 
which provides human-annotated directional queries 
on real-world 360 data, and 
OpenView-VQA~\cite{chen2025openviewempoweringmllmsoutofview}, showing
that our framework is a generalized spherical understanding mechanism and adapts to similar spatial reasoning tasks without any dataset-specific tuning. Finally, Fig.~\ref{fig:teaser-fig}(a) shows a 
qualitative real-world example on an in-the-wild 
360 image, where \textsc{SphMind} correctly 
resolves a directional query that the baseline 
VLM fails to demonstrate that spherical geometric 
grounding transfers beyond controlled benchmarks 
to unconstrained panoramic scenes.

\noindent In summary, our key contributions are:
\textbf{({I})}~A training-free framework that
decouples MLLM semantic perception from geometric reasoning.
\textbf{({II})}~\textbf{SHSG}: a spherical
harmonics scene graph capturing non-Euclidean topology via
continuous SH fields and rotational equivariance.
\textbf{({III})}~\textbf{IGG}: a model-agnostic
latent optimization that enforces geometric constraints on
the MLLM's internal representations at inference time.
\textbf{({IV})}~State-of-the-art gains over MLLM
baselines with strong transferability to similar tasks and
real-world panoramic settings.

\section{Related works}
\begin{wrapfigure}{r}{0.6\linewidth}
  \vspace{-10pt}
  \centering
  \includegraphics[width=\linewidth]{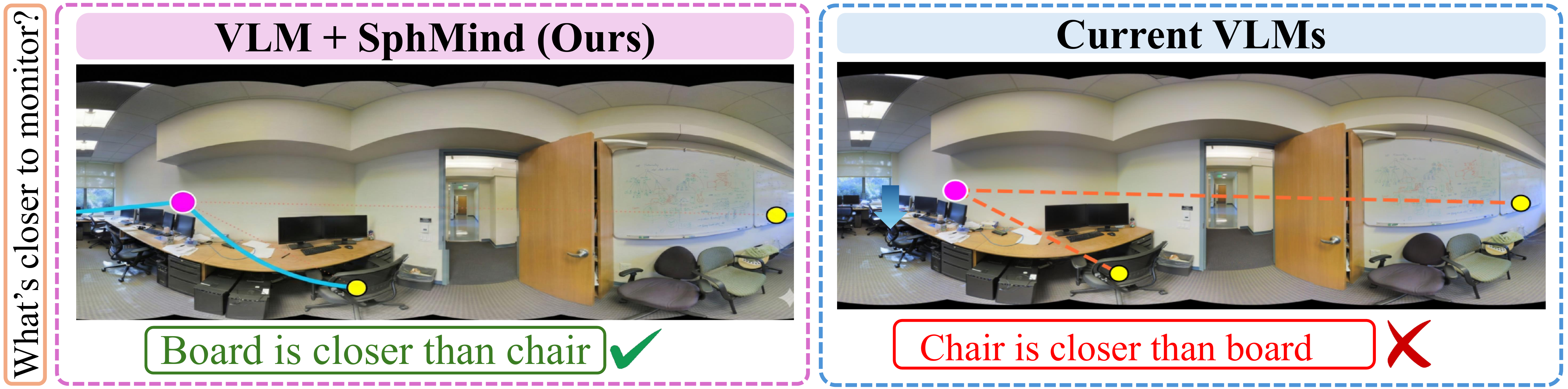}
  \vspace{-12pt}
  \caption{
\textbf{Geodesic distances resolve ERP distortion.}
VLMs misjudge depth from pixel size and wrap-around, ranking the chair closer than the monitor.
\textsc{SphMind} uses geodesic priors on $\mathbb{S}^2$ to correctly identify the board as nearer.
}
  \label{fig:geodesic}
  \vspace{-8pt}
\end{wrapfigure}

\textbf{Omnidirectional Scene Understanding.}
Modern VLMs~\cite{li2023evaluatingobjecthallucinationlarge,bai2023qwentechnicalreport,alayrac2022flamingovisuallanguagemodel} excel at narrow FoV images, their performance degrades significantly on ODIs. The standard ERP introduces latitude-dependent distortions that violate the Euclidean assumptions inherent in common 2D architectures~\cite{wang2018omnidirectionalcnnvisualplace,cohen2018sphericalcnns,he2025positioneuclideanfoundation,ai2024elite360d,ai2024elite360m}. Consequently, rotation-invariant spatial concepts (e.g., left/right, near/far) become highly distorted and difficult to represent reliably on a planar grid. To address these geometric challenges, SoTA  methods typically rely on extensive fine-tuning or specialized architectural designs. Sphere-aware models, such as SphereUFormer~\cite{benny2024sphereuformerushapedtransformerspherical} and Distortion-Aware Transformers~\cite{zhang2022bendingrealitydistortionawaretransformers, ai2025survey,zheng2023both, zheng2024semantics,zheng2023look}, modify internal attention operators to respect spherical topology. Other holistic 3D pipelines leverage frequency-domain representations; for instance, HUSH~\cite{11094938} demonstrates that spherical harmonics provide a robust, geometry-aligned basis for panoramic scene understanding directly on $\mathbb{S}^2$  rather than on distorted planar parameterizations~\cite{dong2023panocontextformerpanoramictotalscene}. 
\textsc However, these existing approaches demand expensive fine-tuning and specialized backbone modifications to adapt. Figs.~\ref{fig:sh_qual} and \ref{fig:geodesic} show that models trained on perspective images apply the same assumptions to ERP inputs, leading to inconsistencies with the true spherical geometry and misinterpretation of directional relations and distances.

\noindent\textbf{Omnidirectional Spatial Reasoning and Benchmarking.}
Early omnidirectional reasoning methods, such as VQA 360$^\circ$~\cite{chou2020visualquestionanswering360deg}, adapted planar pipelines by projecting ODIs into cubemap faces to handle ERP distortion and provide a attention fusion mechanism. While practical, this discrete decomposition fragments the continuous spherical manifold, severely degrading global spatial relationships that depend on uninterrupted 360$^\circ$ context. With the rise of MLLMs, benchmarks including OmniVQA~\cite{zhang2025omnidirectionalreasoning360r1dataset} and Pano-AVQA~\cite{yun2021panoavqagroundedaudiovisualquestion} introduced sphere-aware embeddings to enhance local recognition. Nevertheless, performance drops remain substantial for complex spatial tasks such as relative direction and distance estimation. Recent diagnostic studies~\cite{dongfang2025multimodallargelanguagemodels, lin2026panoenvexploring3dspatial} confirm these persistent weaknesses, revealing a critical \textit{spatial-semantic gap}: MLLMs excel at semantic encoding yet fail to preserve geometry under non-Euclidean topology.
\textit{In contrast, our \textsc{SphMind} introduces a fundamentally different paradigm}.
Rather than forcing a MLLM to implicitly learn spherical geometry through exhaustive retraining, we explicitly decouple semantic perception from geometry.

\noindent \textbf{Spherical Harmonics and Scene Graphs.}
To bridge this gap, structured intermediate representations, such as scene graphs, are employed~\cite{souza2023selfgraphvqaselfsupervisedgraphneural, ainslie2023gqatraininggeneralizedmultiquery}. However, conventional scene graphs are inherently Euclidean and coordinate-dependent. In omnidirectional data, global rotations invalidate discrete directional predicates ( ``left of''), and planar distances become unreliable near projection seams. Works like OSR-Bench \cite{dongfang2025multimodallargelanguagemodels} map the $360^\circ$ image into a 2D ``Cognitive Map'' (an orthographic bird's-eye view) to anchor the MLLM. However, this discretization irreversibly collapses the $Z$-axis, rendering height-based spatial queries unanswerable. More recently, architectures such as Pano-Env \cite{lin2026panoenvexploring3dspatial} and OmniVQA \cite{zhang2025omnidirectionalreasoning360r1dataset} attempt to force spatial awareness via Group Relative Policy Optimization (GRPO), a reinforcement learning paradigm that rewards correct Chain-of-Thought (CoT) reasoning. While effective, training based alignment is computationally expensive and treats spatial laws as stochastic linguistic probabilities rather than immutable physical constants.
Motivated by spherical harmonics as a rotation-aware basis, recent works, such as Equiformer ~\cite{11094938, liao2023equiformerequivariantgraphattention,ai2024dream360}, have tried to enhance panoramic understanding and spatial relationships in omnidirectional images through equivariant geometric learning. This raises a natural question: \textit{given a rotation-consistent spatial state, how can we enforce geometric constraints and correct a frozen MLLM at inference time?}

\noindent \textbf{Adapting to Frozen MLLMs.} Training-free approaches often rely on post-hoc heuristics such as confidence filtering, re-ranking, or localization-based de-hallucination~\cite{chen2025groundsightaugmentingvisionlanguagemodels}, but MLLMs can remain overconfident in physically impossible layouts~\cite{hosseini2026spurlens,HUANG2025109507}, and late-stage filtering cannot repair an internally inconsistent spatial state~\cite{rohrbach2019objecthallucinationimagecaptioning,dziri2023faithfatelimitstransformers}.
Our approach instead aligns with controlled generation and inference-time steering: plug-and-play methods such as PPLM~\cite{dathathri2020plugplaylanguagemodels} show that frozen models can be guided by optimizing latent activations. GrAInS \cite{nguyen2025grainsgradientbasedattributioninferencetime} aims to enable interpretable, token-level control over VLMs via Integrated Gradients to shift from undesirable hallucinations toward truthful responses. Subsequent works show that latent intervention can improve faithfulness and control~\cite{sivakumar2025steervlmrobustmodelcontrol,he2025learningponderadaptivereasoning}.  \textit{Crucially, \textsc{SphMind} extends this paradigm from latent steering to explicit geometric control to realize two advantages}.Unlike prior methods that either fragment the visual space through geometric decompositions or require costly geometry-specific fine-tuning , {\textsc{SphMind} bridges this gap in a training-free manner by reasoning directly on the spherical manifold}. 

\section{Methodology}
\label{sec:method}

\begin{figure*}[t!]
    \centering
    \includegraphics[width=1.0\linewidth]{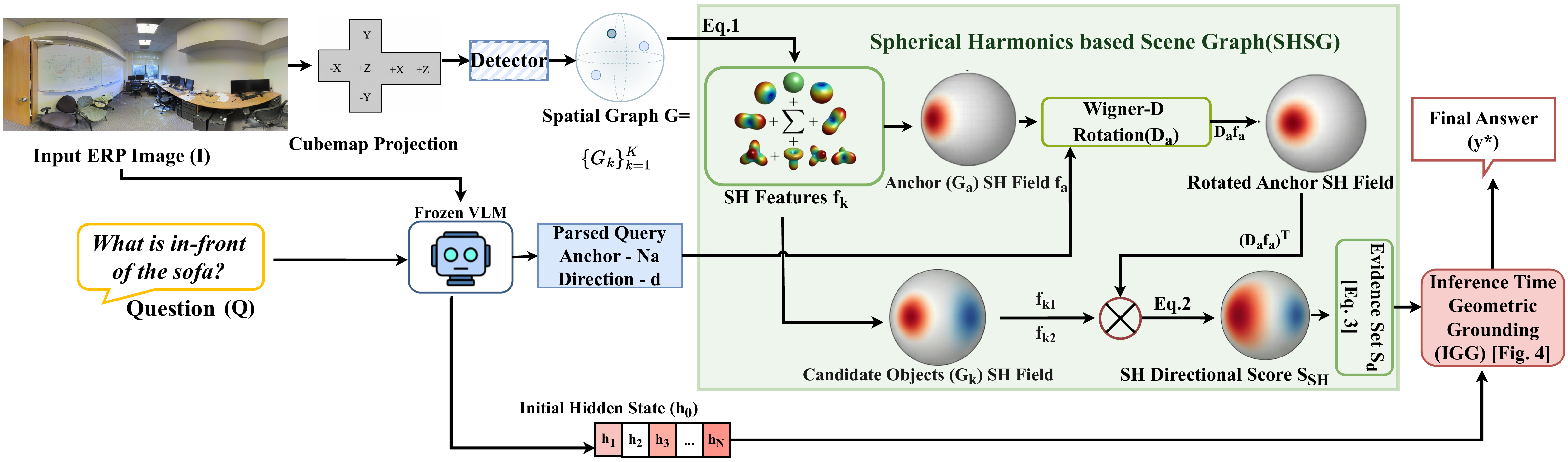}
    \vspace{-13pt}
    \caption{\textbf{Overview of \textsc{SphMind}.}
Given an ERP image and query, SHSG lifts detections to $\mathbb{S}^2$ and builds an SH-encoded scene graph. Wigner-D scoring produces a candidate answer set $\mathbf{Y}$. IGG then steers the frozen VLM hidden state to the final grounded answer $y^*$.}
    \label{fig:method-fig}
    \vspace{-13pt}
\end{figure*} 

\noindent\textbf{Overview.}
\label{sec:overview}
Our central idea is to \emph{decouple semantic
understanding from geometric reasoning}. Put simply, rather than
forcing VLMs to internalize spherical geometry, we
preserve their semantic priors and externalize geometry.
Intuitively, we propose a novel SphMind approach, as shown in Fig. \ref{fig:method-fig}. Given an ERP image ${I}$ and a spatial reasoning question $Q$, 
we seek an answer $y^*$ that is both semantically meaningful and 
geometrically consistent with the 360 scene. Our framework 
achieves this via two tightly coupled novel components. Specifically, \textbf{Spherical Harmonics-based Spatial Graph (SHSG)} 
(Sec.~\ref{sec:shsg}) builds a rotation-consistent scene representation 
directly on $\mathbb{S}^2$ by encoding objects and their spatial 
relations as spherical harmonic features, producing  $\mathbf{Y}$ , a set of geometrically plausible VQA answers for 
$y^*$. However, a pretrained VLM operating on ${I}$ and $Q$ 
remains unaware of how to understand this geometric signal and correct its internal representation. 
\textbf{Inference Time Geometric Grounding (IGG)} (Sec.~\ref{sec:igg}) bridges this gap by injecting the SHSG 
geometric constraints directly into the VLM's final hidden state 
$\mathbf{h}_0$. Thereby, IGG steers this $\mathbf{h}_0$ towards geometrically grounded distribution while preserving the semantic priors. 
In short, the VLM answers \emph{what} is present; SHSG and IGG enforce the VLM on
\emph{where} and \emph{how} the objects relate on the sphere. Now we describe these two technical components in details.

\subsection{Spherical Harmonics-based Spatial Graph (SHSG) Module}
\label{sec:shsg}

As depicted in Fig.~\ref{sec:method}, given an ERP image ${I}$ and a spatial query $Q$, 
we localize objects by projecting ${I}$ into 
cubemap faces and lifting each raw detection outputs 
$\tilde{G}_k$ obtained from object detectors \cite{redmon2016lookonceunifiedrealtime} to spherical coordinates 
$(\theta_k, \phi_k)$ via the corresponding face 
rotation matrix $R_i \in \mathrm{SO}(3)$. Here, each 
node $G_k = (c_k,\,\theta_k,\,\phi_k,\,\hat{c}_k)$ 
carries the object's semantic category $c_k$, 
viewer-centric spherical coordinates $(\theta_k,\phi_k)$, 
and detector confidence $\hat{c}_k \in [0,1]$. Note that cross-face 
duplicates are removed via \emph{geodesic 
non-maximum suppression} (GNMS; 
Appendix~\ref{app:gnms}), yielding a duplicate-free 
spatial graph $\mathbf{G} = \{G_k\}_{k=1}^{K}$ where $K$ is the total number of nodes.  The 
spatial graph $\mathbf{G}$ and query $Q$, 
specifying an anchor object node $G_a$ and a target 
direction $d$, together form the inputs to SHSG.

Consequently, a fundamental requirement for omnidirectional reasoning 
is \emph{rotational consistency}: spatial relations 
between objects must transform predictably when the 
observer rotates. Standard Euclidean representations 
break under such rotations because translation-invariant 
mechanisms cannot model the periodicity or distortions 
of the sphere. We address this by representing 
each detected object node $G_k$ directly on the sphere 
using real \textbf{spherical harmonic (SH) node 
encoding} \cite{dai2013sphericalharmonics}. Spherical harmonics $Y_{\ell}^{m}(\theta, \phi)$ 
form a 
complete orthonormal basis on the sphere
and transform \emph{linearly} 
and \emph{exactly} under $\mathrm{SO}(3)$ rotations. 
The SH feature of object node $G_k$ is written as:
\begin{equation}
    \mathbf{f}_k
    =
    \bigl[Y_{\ell}^{m}(\theta_k, \phi_k)
    \bigr]_{\substack{\ell=0,\ldots,L_{\max}\\
                      m=-\ell,\ldots,\ell}}
    \in \mathbb{R}^{(L_{\max}+1)^2}.
    \label{eq:sh_feat}
\end{equation}
For $L_{\max}{=}7$, this yields a 64-dimensional 
rotation-equivariant encoding of the object's angular 
position, sufficient to resolve 
spatial features at ${\approx}20^\circ$ resolution 
(see Appendix~\ref{app:why_sh}). As shown in Fig. \ref{fig:method-fig}, we consider any two object nodes, $G_1$ and  $G_2$ and their SH-encoded feature as $\mathbf{f}_{k,1}$ and $\mathbf{f}_{k,2}$ while the anchor object node's feature is represented as $\mathbf{f}_{a}$.
This
contains all geometric information about the angular 
position of object $G_k$ that is relevant for 
directional reasoning from a fixed viewpoint.

\noindent\textbf{Directional Reasoning }
Given anchor object node $G_a$ and query direction $d$, 
SHSG answers \emph{``which object lies in direction 
$d$ from $G_a$?''} directly on the sphere $\mathbb{S}^2$, without 
reference to the distorted ERP pixel space. In 
spherical geometry, this question is equivalent to 
rotating the anchor object's coordinate frame by the 
rotation $R_d \in \mathrm{SO}(3)$ that maps the 
canonical forward direction to $d$, then identifying 
which candidate object $G_k$ aligns most strongly with 
the rotated anchor. Because each object is encoded as 
an SH feature ${f}_k$, and SH features transform 
linearly under $\mathrm{SO}(3)$ via the 
Wigner-$\mathcal{D}$ matrix, this rotation is exact 
and closed-form. The \textbf{SH directional score} $\mathrm{SSH}$
measuring how well candidate object $G_k$ satisfies 
the directional constraint relative to anchor $G_a$ is:
\begin{equation}
    S_{\mathrm{SH}}(a,k,d)
    =
    \frac{
        \operatorname{ReLU}\!\bigl(
            (\mathcal{D}_d\,\mathbf{f}_a)^\top
            \mathbf{f}_k
        \bigr)^{\!\gamma}
    }{
        \displaystyle\sum_{k'}
        \operatorname{ReLU}\!\bigl(
            (\mathcal{D}_d\,\mathbf{f}_a)^\top
            \mathbf{f}_{k'}
        \bigr)^{\!\gamma}
        + \varepsilon
    },
    \label{eq:sh_score}
\end{equation}
where $\mathcal{D}_d$ is the Wigner-$\mathcal{D}$ 
operator and
$\varepsilon{=}10^{-8}$ 
ensures numerical stability. The rotated anchor feature 
$\mathcal{D}_d\,\mathbf{f}_a$ encodes what direction 
$d$ from $G_a$ looks like in SH space. The inner product \((\mathcal{D}_d\,\mathbf{f}_a)^\top \mathbf{f}_k\) is positive when \(G_k\) aligns with direction \(d\) from \(G_a\), and negative for antipodal directions. A \(\operatorname{ReLU}\) enforces this sign structure, suppressing antipodal candidates in a parameter-free manner.
\begin{wrapfigure}{r}{0.68\linewidth}
  \centering
  \vspace{-5pt}
  \includegraphics[width=\linewidth]{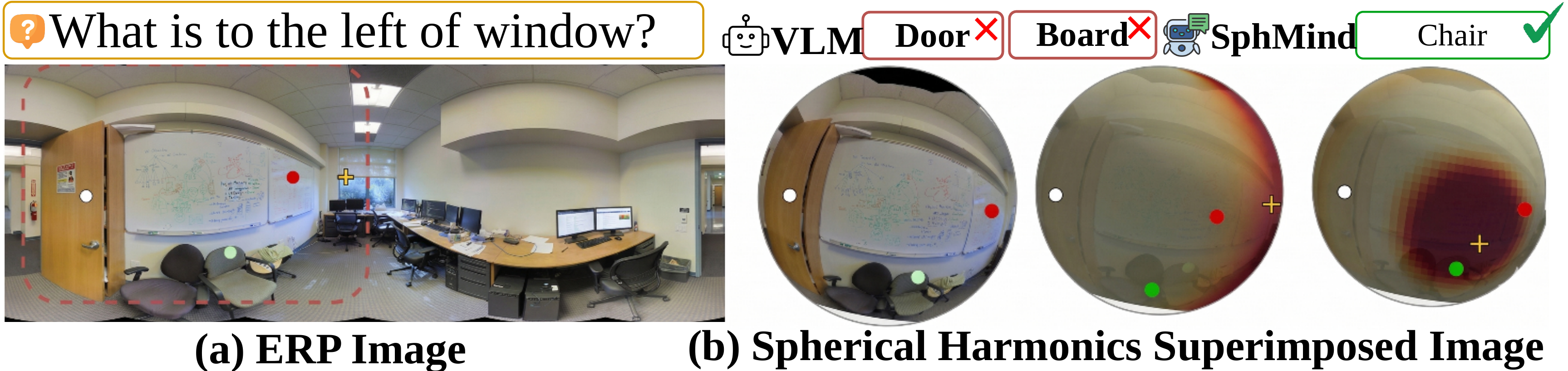}
  \vspace{-10pt}
  \caption{\textbf{SHSG directional reasoning on MP3D.}
  Query: \textit{``What is to the left of the window?''}
  \textbf{(a)}~ERP image with candidates:
  \textcolor{gray}{$\bullet$}~door,
  \textcolor{red}{$\bullet$}~board,
  \textcolor{green}{$\bullet$}~chair,
  $\boldsymbol{+}$~window anchor.
  \textbf{(b)}~SH spherical views: RGB scene on
  $\mathbb{S}^2$ (left); SH field centred at
  anchor (center); after Wigner-D rotation the peak
  shifts left, aligning with \textcolor{green}{$\bullet$}~chair
  (right).}
  \label{fig:sh_sphere}
    \vspace{-10pt}
\end{wrapfigure}
Further, \(S_{\mathrm{SH}}\) is a query-conditioned operator on the sphere, where the Wigner-\(\mathcal{D}_d\) rotation encodes direction \(d\) and produces distinct scores for different directional queries over the same candidates. It is equivariant to global rotations \(R \in \mathrm{SO}(3)\), ensuring that scores transform consistently and the correct answer is preserved regardless of panorama orientation (refer to Appendix~\ref{app:why_sh}).

The candidate object nodes $\mathbf{N}$ are geometrically consistent with direction 
$d$ from anchor object $G_a$ are collected as the 
\textbf{directional evidence set}:
\begin{equation}
    \mathbf{S}_d
    =
    \bigl\{
        \bigl(G_k,\;S_{\mathrm{SH}}(a,k,d)\bigr)
        \;:\;
        S_{\mathrm{SH}}(a,k,d) > 0
    \bigr\},
    \label{eq:cand}
\end{equation}
written compactly as 
$\mathbf{S}_d = \Psi(\mathbf{G};\,N_a,\,d)$, and 
illustrated on a real Matterport3D scene in 
Fig.~\ref{fig:sh_sphere}. Then, $\mathbf{S}_d$ 
 is passed to 
IGG, which lifts the SH scores into a geometric 
violation cost and 
steers the frozen VLM's hidden state toward the 
geometrically correct answer. We now describe the details.

\noindent\textbf{Distance queries.}
For distance queries, SHSG derives a different scoring mechanism using
camera-frame depth score $s_k$ for each 
candidate from $\mathbb{S}^2$-native signals 
and defines $C(y_A){=}s_B/(s_A{+}s_B)$, 
$C(y_B){=}s_A/(s_A{+}s_B)$, rewarding the 
closer candidate with lower cost.

The depth 
estimation pipeline slightly differs from directional 
scoring and not the main focus.  
(Further details in Appendix~\ref{app:distance}.)

\subsection{Inference-Time Geometric Grounding (IGG)}
\label{sec:igg}

\begin{wrapfigure}{r}{0.7\linewidth}
    \centering
    \vspace{5pt}
    \includegraphics[width=\linewidth]{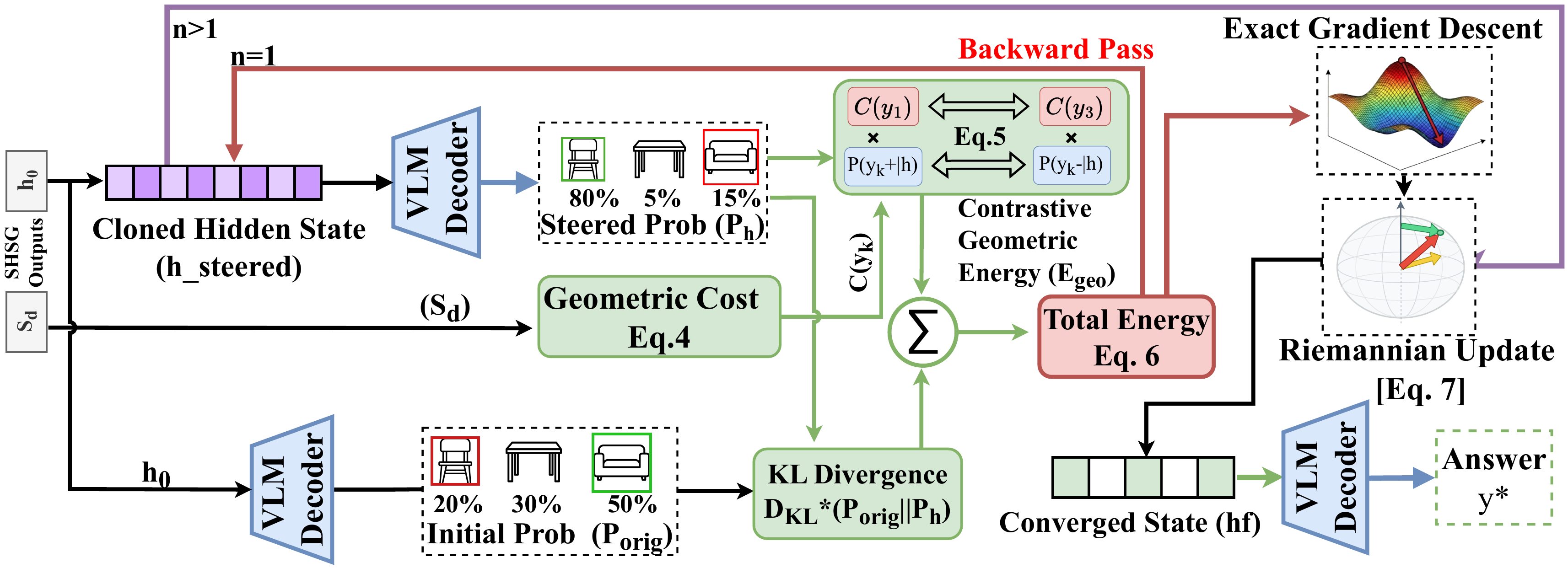}
    \caption{IGG pipeline. Given the VLM hidden state 
    $\mathbf{h}_0$, ranked evidence $\mathbf{S}_d$, IGG performs $T$ norm-preserving Riemannian 
    updates to steer $\mathbf{h}$ toward the 
    geometrically consistent answer $y^*$}
    \label{fig:igg-fig}
    \vspace{-10pt}
\end{wrapfigure}

SHSG is a purely geometric module: it operates entirely on the sphere
and produces a directional evidence set $\mathbf{S}_d$ with SH-based 
geometric scores with no semantic knowledge. However, the frozen VLM's hidden state $\mathbf{h}_0$ encodes rich 
semantic priors but carries no spherical understanding. Therefore, our IGG bridges this gap by propagating SHSG's geometric evidence directly into 
$\mathbf{h}_0$ via energy minimisation, as shown in Fig.\ref{fig:igg-fig}. That is, IGG aims to ground the VLM's output 
distribution in the spherical geometry of the scene prior to 
decoding without updating any model weights.

\noindent\textbf{Geometric Violation Cost.}
Given the directional evidence set 
$\mathbf{S}_d$ 
with SH directional scores $S_{\mathrm{SH}}(a,k,d)$ 
for each candidate object node $G_k$, where a higher 
score reflects stronger directional alignment, we define the geometric violation cost as the 
complement of the SH score:
\begin{equation}
    C(y_k) = 1 - S_{\mathrm{SH}}(a,k,d),
    \quad y_k \in \mathbf{S}_d.
    \label{eq:cost}
\end{equation}
This assigns each candidate in Sd 
a geometric violation cost inversely proportional 
to its directional alignment, providing the 
structured geometric signal for IGG.
While the cost $C(y_k)$ provides a geometric 
penalty for each candidate token in $\mathbf{S}_d$, 
the VLM's likelihood over these tokens is governed 
entirely by its continuous latent representation 
$\mathbf{h}_0 \in \mathbb{R}^d$ , the final hidden 
state after processing image $\mathbf{I}$ and query 
$Q$. Since $\mathbf{h}_0$ is shaped by pretrained backbones with no awareness of spherical 
geometry, we inject the geometric signal from SHSG 
directly into this latent space via energy 
minimisation.

\vspace{-3pt}
A pointwise expected violation 
$\sum_k P(y_k|\mathbf{h})\,C(y_k)$ produces a 
gradient proportional to $P(y_k|\mathbf{h})$ , 
weak precisely when the VLM assigns low probability 
to the geometrically correct answer and high 
probability to wrong candidates, i.e.\ when 
correction is most needed. We instead define the 
\textbf{Contrastive geometric energy} 
$E_{\mathrm{geo}}$ as a pairwise ranking objective 
over all geometrically ordered pairs in 
$\mathbf{S}_d$:
\begin{equation}
    E_{\mathrm{geo}}(\mathbf{h})
    =
    \frac{1}{|\mathcal{P}|}
    \sum_{(k^+,\,k^-)\in\mathcal{P}}
    \bigl[C(y_{k^-}) - C(y_{k^+})\bigr]
    \log\frac{P(y_{k^-}\mid\mathbf{h})}
             {P(y_{k^+}\mid\mathbf{h})},
    \label{eq:egeo}
\end{equation}
where $\mathcal{P} = \{(k^+,k^-) : C(y_{k^-}) > 
C(y_{k^+}),\; k^\pm \in \mathbf{S}_d\}$ is the 
set of all geometrically ordered pairs, $k^+$ the 
more directionally consistent candidate and $k^-$ 
the less consistent. The total energy is:
\begin{equation}
    E(\mathbf{h})
    =
    \beta_{\mathrm{geo}}\,E_{\mathrm{geo}}(\mathbf{h})
    +
    \beta_{\mathrm{KL}}\,
    D_{\mathrm{KL}}\!\bigl(
        P_{\mathrm{orig}} \,\|\, P_{\mathbf{h}}
    \bigr),
    \label{eq:energy}
\end{equation}
where $P_{\mathrm{orig}} = P(y\mid\mathbf{h}_0)$ is 
the original VLM distribution. $E_{\mathrm{geo}}$ 
penalises every pair where the VLM assigns higher 
probability to the geometrically wrong candidate 
than the correct one, with the penalty scaled by 
both how geometrically discriminative the pair is 
$[C(y_{k^-}){-}C(y_{k^+})]$ and how severely the 
probability ordering is violated 
$[\log P(y_{k^-}|\mathbf{h})/P(y_{k^+}|\mathbf{h})]$ 
— producing the strongest corrective gradient 
precisely when it is most needed. The KL term 
prevents the steered distribution from collapsing 
onto a single token by anchoring it to the VLM's 
pretrained semantic prior. 
Together they realise a \emph{geometrically-ordered 
semantic optimisation}:  which aligns the VLM's probability 
ordering with the geometric ranking from SHSG while 
remaining close to the pretrained prior. 
$E_{\mathrm{geo}}$ reaches zero when 
$P_{\mathbf{h}}$ is monotonically decreasing in 
$C(y_k)$ meaning the VLM's distribution perfectly 
reflects the geometric ranking 
(refer to Appendix~\ref{app:energy_zero}).

\noindent\textbf{Riemannian Update.}
The gradient $\mathbf{g} = \nabla_{\mathbf{h}} 
E(\mathbf{h})$, obtained by differentiating 
through the frozen LM head $\mathbf{W}_{\mathrm{lm}}$, 
is a weighted sum of differences of LM head rows 
corresponding to geometrically ordered pairs , 
propagating the geometric ranking through the 
VLM's own inter-token semantic structure without 
any weight update (refer to Appendix~\ref{app:grad_proof}). 
Since $\mathbf{h}$ lies on the $\ell_2$-norm 
hypersphere $\mathcal{M} = \{\mathbf{h} : 
\|\mathbf{h}\| = r\}$, a naive Euclidean step 
would change $\|\mathbf{h}\|$ and alter the 
decoding temperature via LayerNorm. We therefore 
project $\mathbf{g}$ onto the tangent space 
$T_{\mathbf{h}}\mathcal{M}$ and retract back 
onto $\mathcal{M}$:
\begin{equation}
    \mathbf{h}_{t+1}
    =
    r \cdot
    \frac{
        \mathbf{h}_t - \eta\,
        \Bigl(
            \mathbf{g}_t
            -
            \frac{\mathbf{g}_t^\top \mathbf{h}_t}
                 {\|\mathbf{h}_t\|^2}
            \mathbf{h}_t
        \Bigr)
    }{
        \Bigl\|
            \mathbf{h}_t - \eta\,
            \Bigl(
                \mathbf{g}_t
                -
                \frac{\mathbf{g}_t^\top \mathbf{h}_t}
                     {\|\mathbf{h}_t\|^2}
                \mathbf{h}_t
            \Bigr)
        \Bigr\|
    },
    \label{eq:riemannian_update}
\end{equation}
where $r = \|\mathbf{h}_0\|$ is preserved exactly 
at every step, isolating the geometric correction 
as a pure directional intervention 
(refer to Appendix~\ref{app:riemannian_proof}). The 
contrastive gradient is self-scaling with the 
degree of geometric ordering violation , strong 
when the VLM is most miscalibrated, vanishing 
when the probability ordering already matches 
the geometric ranking  making a single 
Riemannian step sufficient for the majority of 
cases without requiring empirical step-count tuning.

\noindent\textbf{Final Prediction.}
After a single Riemannian step with early stopping 
when $\|\Delta E\| < \delta$ 
(refer to Appendix~\ref{app:stepsize}), we obtain the 
geometrically grounded hidden state $\mathbf{h}'$. 
The final answer is decoded by restricting to the 
directionally consistent candidate set:
\(y^* = \arg\max_{y_k \in \mathbf{S}_d} 
P(y_k \mid \mathbf{h}')\).
The VLM focuses on \emph{what} objects are 
semantically present; SHSG and IGG together 
enforce \emph{where} they are geometrically 
on the sphere $\mathbb{S}^2$.

\begin{figure}[t!]
  \centering
  \vspace{-10pt}
  \includegraphics[width=\linewidth]{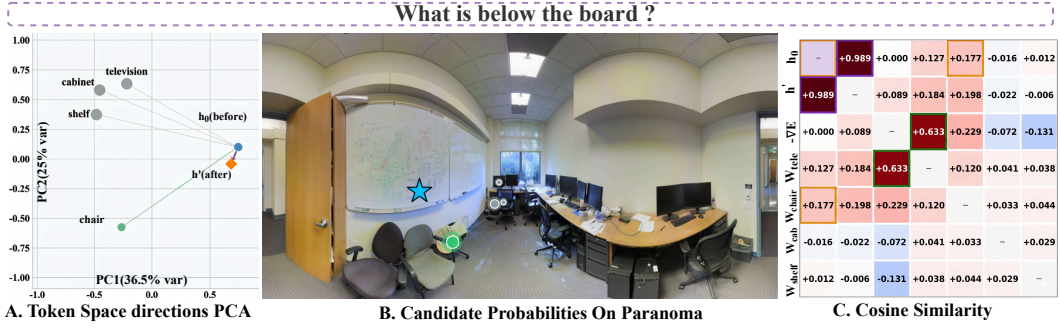}

\caption{
\textbf{(A)}~Token-space PCA: IGG shifts $\mathbf{h}_0 \rightarrow \mathbf{h}'$ toward the correct token (\textcolor{green}{chair}) while preserving semantic structure.  
\textbf{(B)}~Panorama candidates: \textcolor{cyan}{$\bigstar$} marks the board; \textcolor{green}{$\bullet$} indicates the correctly localized chair below it on $\mathbb{S}^2$.  
\textbf{(C)}~Cosine similarity: confirms semantic consistency, valid tangent update, and correct gradient alignment, showing correction of the initial model bias through geometric grounding.}
  \vspace{-10pt}
  \label{fig:matrix}
\end{figure}

\begin{table}[t!]
\centering
\caption{Performance across datasets and scenes.}
\label{tab:scene_results}
\small
\setlength{\tabcolsep}{3pt}
\renewcommand{\arraystretch}{1.05}

\resizebox{\columnwidth}{!}{
\begin{tabular}{cc|l|cccccc|cccc}
\toprule
\textbf{Model} & \textbf{Type} & \textbf{Category}
& \multicolumn{6}{c|}{\textbf{Matterport3D}}
& \multicolumn{4}{c}{\textbf{Stanford2D}} \\

& & & Scene 1 & Scene 2 & Scene 3 & Scene 4 & Scene 5 & Scene 6
& Scene 1 & Scene 2 & Scene 3 & Scene 4 \\
\midrule

\multirow{10}{*}{\rotatebox[origin=c]{90}{\textbf{Qwen2.5-7B}}}
& \multirow{5}{*}{Baseline}
& Overall   & 40  & 42.9 & 36.2 & 24.3 & 41.7 & 38.6 & 35.7 & 39.8 & 31.5 & 37.2 \\
& & Presence  & 62.5 & 67.8 & 62.5 & 53.5 & 72.2 & 64.5 & 100 & 70.3 & 58.4 & 66.9 \\
& & Counting  & 20 & 10 & 14.2 & 13 & 11.1 & 20 & 8.3 & 17.6 & 9.8 & 15.2 \\
& & Distance  & 40 & 40 & 27 & 42.2 & 44.8 & 41.7 & 50 & 45.6 & 34.9 & 43.8 \\
& & Direction & 35.3 & 41.2 & 28.6 & 22.5 & 32.1 & 20.7 & 14.3 & 32.8 & 24.7 & 30.4 \\
\cline{2-13}
\addlinespace[2pt]
& \multirow{5}{*}{\textbf{Ours}}
&  Overall   & 65 & 63.3 & 56.8 & 50.8 & 52.4 & 59.6 & 62.5 & 60.7 & 54.1 & 58.9 \\
& & Presence  & 80 & 75 & 65.7 & 59.8 & 66.7 & 72.4 & 93.3 & 78.6 & 63.2 & 74.9 \\
& & Counting  & 50.6 & 47.5 & 33.6 & 24.5 & 26.1 & 38.7 & 25 & 36.4 & 27.3 & 34.8 \\
& & Distance  & 83.3 & 71.6 & 72.5 & 53.3 & 60 & 61.2 & 55.7 & 51.5 & 62.4 & 69.8 \\
& & Direction & 58.3 & 55.2 & 54.3 & 52.6 & 51.1 & 53.8 & 50 & 54.7 & 48.9 & 52.1 \\
\midrule
\addlinespace[2pt]

\multirow{10}{*}{\rotatebox[origin=c]{90}{\textbf{LLaVA-1.5-7B}}}
& \multirow{5}{*}{Baseline}
& Overall   & 38.3 & 35.7 & 34.1 & 36.5 & 33.8 & 37.2 & 32.1 & 45.6 & 28.9 & 36.7 \\
& & Presence  & 68.8 & 64.2 & 66.1 & 65.7 & 63.9 & 67.4 & 67.2 & 69.4 & 68.1 & 70.5 \\
& & Counting  & 20.0 & 18.7 & 19.5 & 17.9 & 16.8 & 21.4 & 18.5 & 22.3 & 19.8 & 21.0 \\
& & Distance  & 40.0 & 42.6 & 38.4 & 39.7 & 31.3 & 33.5 & 35.4 & 48.7 & 42.1 & 49.5 \\
& & Direction & 35.3 & 30.4 & 32.1 & 29.6 & 31.7 & 34.2 & 28.5 & 35.9 & 26.8 & 33.1 \\
\cline{2-13}
\addlinespace[2pt]
& \multirow{5}{*}{\textbf{Ours}}
& Overall   & 60.9 & 57.8 & 58.6 & 59.2 & 56.7 & 60.1 & 58.2 & 63.4 & 60.1 & 61.6 \\
& & Presence  & 90.0 & 88.7 & 89.3 & 87.9 & 88.5 & 90.6 & 89.2 & 90.8 & 89.7 & 91.2 \\
& & Counting  & 23.5 & 22.8 & 24.1 & 21.9 & 22.6 & 24.3 & 22.1 & 24.9 & 23.8 & 24.2 \\
& & Distance  & 69.6 & 73.4 & 67.1 & 74.2 & 66.3 & 68.5 & 70.3 & 71.9 & 72.8 & 77.4 \\
& & Direction & 48.8 & 45.6 & 47.2 & 44.9 & 46.8 & 49.4 & 46.4 & 51.2 & 47.9 & 50.5 \\
\midrule
\addlinespace[2pt]

\multirow{10}{*}{\rotatebox[origin=c]{90}{\textbf{Qwen2.5-3B}}}
& \multirow{5}{*}{Baseline}
& Overall   & 35.8 & 38.6 & 33.4 & 29.7 & 31.2 & 34.1 & 42.7 & 37.9 & 30.5 & 36.4 \\
& & Presence  & 85.0 & 88.4 & 84.2 & 81.3 & 87.1 & 89.6 & 93.0 & 86.7 & 82.5 & 88.9 \\
& & Counting  & 14.0 & 12.5 & 15.2 & 13.8 & 10.9 & 18.3 & 16.7 & 14.6 & 12.8 & 15.1 \\
& & Distance  & 41.2 & 44.7 & 38.6 & 35.9 & 39.4 & 46.8 & 50.0 & 43.5 & 37.8 & 45.2 \\
& & Direction & 9.5 & 12.3 & 14.8 & 11.7 & 11.4 & 13.2 & 17.5 & 11.1 & 10.2 & 12.0 \\
\cline{2-13}
\addlinespace[2pt]
& \multirow{5}{*}{\textbf{\textbf{Ours}}}
& Overall   & 54.6 & 56.3 & 52.8 & 49.1 & 50.7 & 53.2 & 58.9 & 55.4 & 51.6 & 57.2 \\
& & Presence  & 93.8 & 95.1 & 92.4 & 91.3 & 94.2 & 96.0 & 96.3 & 94.5 & 92.1 & 95.8 \\
& & Counting  & 22.4 & 23.6 & 24.8 & 21.7 & 20.9 & 26.4 & 25.0 & 23.1 & 21.5 & 24.3 \\
& & Distance  & 68.9 & 71.5 & 69.7 & 65.2 & 67.4 & 72.1 & 74.0 & 70.6 & 66.3 & 72.8 \\
& & Direction & 36.2 & 38.4 & 39.5 & 35.8 & 37.6 & 41.3 & 40.5 & 38.9 & 36.7 & 41.0 \\
\bottomrule
\end{tabular}
}
\vspace{-10pt}
\end{table}

\section{Experiments}
\label{sec:experiments}
\subsection{Experimental Setup}

\noindent\textbf{Datasets.}
We evaluate \textsc{SphMind} on three benchmarks:
\textit{(I) Matterport3D (MP3D) and Stanford 2D-3D-S}:
$\sim$12k synthetic QA pairs derived from 3D annotations provided
across six MP3D scenes and four 2D-3D-S areas.
\textit{(II) ODI-Bench}: $\sim$1k human-annotated QA pairs
on real-world Flickr ODIs; we test zero-shot on
Relative Direction and Egocentric View Orientation.
\textit{(III) OpenView-VQA}: $\sim$1k human-annotated camera-rotation
reasoning QA pairs from restricted FoV crops.

\noindent\textbf{Benchmark Construction.}
For MP3D and 2D-3D-S, ground-truth answers of VQA are derived from the provided annotations i.e 3D object centroids, semantic labels,
and camera poses, projected to viewer-centric spherical
coordinates $(\theta,\phi,r)$ (refer to Appendix~\ref{app:benchmark}).
ODI-Bench and OpenView-VQA provide human-annotated QA pairs
with raw panoramic images and camera poses.

\noindent\textbf{Baselines.} We compare \textsc{SphMind} across three open-source VLM backbones spanning different architectures and scales: \textbf{Qwen2.5-VL} (3B, 7B) and \textbf{LLaVA-1.6} (7B). Each backbone is evaluated under two settings: \emph{Vanilla}, where the frozen VLM receives the raw ERP panorama and question, and \emph{+SphMind}, where the same frozen VLM is augmented with SHSG+IGG. Presence and counting are evaluated without external detections to measure the raw ERP perception gap; direction and distance use identical YOLO detections and graph $\mathbf{G}$ for both settings, isolating SHSG+IGG as the sole variable (see Appendix~\ref{app:eval_protocol} for full protocol details).

\vspace{-10pt}
\subsection{Results and Discussion}
\vspace{-5pt}
We structure our evaluation around the following experimental questions (EQs):

\vspace{-5pt}
\begin{center}
\colorbox{gray!10}{
\begin{minipage}{0.96\linewidth}

\textbf{EQ-1:} Can explicit spherical geometric grounding enable frozen VLMs to
reason about panoramic spatial relationships that they fundamentally
cannot understand in ERP images?

\textbf{EQ-2:} Does the geometric prior in \textsc{SphMind} generalize
zero-shot across structurally distinct benchmarks, invariant to rotations and coordinate conventions without any task-specific training?

\textbf{EQ-3:} Which components of \textsc{SphMind} are necessary for the observed
gains in spatial reasoning, and  is the $\mathbb{S}^2$ spherical representation necessary for panoramic understanding 

\end{minipage}
}
\end{center}

\vspace{-4pt}
\noindent\textbf{360\textdegree\ Panoramic VQA.}
Table~\ref{tab:scene_results} reveals that direction baselines expose the most
severe failure: Qwen2.5-3B reaches only $9.5$--$17.5\%$ across scenes,
confirming misinterpretation of directions in pixel space.
Distance baselines cluster at $35$--$50\%$, revealing that VLMs actively misread ERP-distortions. Presence and counting baselines are substantially
higher, showing the failure is specific to geometry-intensive queries
rather than general VLM weakness. Gains are consistent across
architectures and scale, ruling out model-specific adaptation.

\begin{center}
\vspace{-3pt}
\colorbox{gray!10}{
\begin{minipage}{0.96\linewidth}
\textbf{[A-1]} \textsc{SphMind} raises overall accuracy from
$35.9\%$ to $56.4\%$ \textbf{($+20.5\%$)}. 
Direction gains are largest for weakest model (Qwen2.5-3B: $\sim$$12\%{\to}\sim$$39\%$) and smallest for strongest (LLaVA-1.5-7B: $\sim$$31\%{\to}
\sim$$48\%$) proving increased geometric understanding
 rather
than a semantic bias being corrected.
Distance is consistently around $60\%$
for all models and scenes. 
\end{minipage}}
\vspace{-3pt}
\end{center}

\vspace{-2pt}
\noindent\textbf{Zero-Shot Transfer to ODI-Bench.}
Table~\ref{tab:combined}~(a) compares \textsc{SphMind} against four
prompt-engineering strategies on real-world Flickr ODIs.
All methods remain within $\pm 1.5\%$ of the $46.1\%$ RD baseline, with crop
grounding actually degrading to $44.2\%$, confirming that
linguistic or input-space interventions cannot substitute for incorrect
geometric understanding.

\vspace{-2pt}
\noindent\textbf{Rotation Angle Grounding on OpenView-VQA.}
Table~\ref{tab:combined}(b) tests whether exact 
spherical conversion of camera rotation angles 
outperforms language based rotation 
interpretation. OpenView-VQA 
expresses spatial targets via rotation angles 
from restricted FoV crops; a frozen VLM 
associates these with candidates via language 
statistics rather than geometric conversion. 
SHSG converts the rotation angle directly to 
$(\theta_{\mathrm{tgt}}, \phi_{\mathrm{tgt}}) = 
(\theta_{\mathrm{cam}} + \Delta\psi,\; 
\phi_{\mathrm{cam}} + \Delta\rho)$ and substitutes 
it as the anchor, with all other operations 
unchanged.
(refer to Appendix~\ref{app:openview}).

\noindent\textbf{Invariance across Rotated Panorama.}
Table~\ref{tab:combined}(c) measures whether \textsc{SphMind} gives
the same answer under $90^\circ$, $180^\circ$, and $270^\circ$
azimuthal rotations of the ERP origin. ERP based baseline reasoning achieves
only $4.3\%$ full RCS and degrades most severely at $180^\circ$
(RCS $= 0.212$), where front$\leftrightarrow$behind swaps scatter
the relevant scene region across the seam. \textsc{SphMind} achieves
$59.5\%$ at $180^\circ$ as compared to $21.2\%$
for the baseline.

\begin{center}
\vspace{-5pt}
\colorbox{gray!10}{
\begin{minipage}{0.96\linewidth}
\textbf{[A-2]} 
\textsc{SphMind} reaches $56.35\%$ RD and $52.60\%$ EVO on
ODI-Bench, outperforming all prompt-engineering strategies by at
least $+8.7\%$ on RD. On OpenView-VQA, it achieves $39.7\%$
 $21.0\%$ against the zero-shot baseline, and comparable to method ($46.0\%$) without any task-specific
training. Full RCS reaches $25.3\%$ against $4.3\%$ for ERP-pixel
reasoning ($\times5.9$), with the largest gain at $180^\circ$
($+0.383$) where pixel reasoning is most structurally broken.
\end{minipage}}
\vspace{-5pt}
\end{center}

\begin{table*}[t!]
\centering
\footnotesize
\setlength{\tabcolsep}{3pt}
\renewcommand{\arraystretch}{1.0}

\begin{minipage}[t]{0.31\linewidth}
\centering
\scriptsize
\begin{tabular}{lcc}
\toprule
\textbf{Strategy} & \textbf{RD} & \textbf{EVO} \\
\midrule
Baseline              & 46.13 & 45.71 \\
+ Viewpoint Guidance  & 45.80 & 45.10 \\
+ Crop Grounding      & 44.18 & 44.15 \\
+ Response Refinement & 47.65 & 45.76 \\
+ CoT                 & 46.33 & 43.20 \\
\midrule
\textbf{+ SphMind}    & \textbf{56.35} & \textbf{52.60} \\
\bottomrule
\end{tabular}
\vspace{-92pt}
\captionof*{table}{\footnotesize \textbf{(a)}}
\end{minipage}
\hfill
\begin{minipage}[t]{0.28\linewidth}
\centering
\footnotesize
\begin{tabular}{lccc}
\toprule
\textbf{Category} & \textbf{Base} & \textbf{\textbf{Ours}} & \textbf{Train} \\
\midrule
Overall & 21.0 & 39.7 & 46.0 \\
Pitch   & 22.5 & 36.2 & 44.3 \\
Yaw     & 25.0 & 45.6 & 48.7 \\
P+Y     & 23.1 & 42.1 & 45.0 \\
\bottomrule
\end{tabular}
\vspace{-80pt}
\captionof*{table}{\footnotesize \textbf{(b)}}
\end{minipage}
\hfill
\begin{minipage}[t]{0.36\linewidth}
\centering
\footnotesize
\begin{tabular}{lcccc}
\toprule
\textbf{Method} & \textbf{Full} & \textbf{90$^\circ$} & \textbf{180$^\circ$} & \textbf{270$^\circ$} \\
\midrule
ERP-Pixel        & 0.043 & 0.437 & 0.212 & 0.451 \\
\textbf{SphMind} & \textbf{0.253} & \textbf{0.653} & \textbf{0.595} & \textbf{0.648} \\
\midrule
\textit{Gain}    & $\times$5.9 & +0.216 & +0.383 & +0.197 \\
\bottomrule
\end{tabular}
\vspace{-75pt}
\captionof*{table}{\footnotesize \textbf{(c)}}
\end{minipage}

\vspace{35pt}
\caption{\textbf{Comparison across panoramic reasoning benchmarks.}
SphMind consistently improves performance across diverse settings, including \textbf{(a)} object-directional inference (ODI-Bench), \textbf{(b)} viewpoint generalization (OpenView), and \textbf{(c)} rotation-consistent spatial grounding (RCS).}
\label{tab:combined}
\vspace{-15pt}
\end{table*}

\subsection{Ablation Study}
\label{sec:ablation}

\noindent\textbf{Component Analysis.}
Table~\ref{tab:ablation} ablates each component on MP3D
(Qwen2.5-7B); per-category drops reveal which components
are critical for which reasoning type.

\begin{wraptable}{r}{0.45\columnwidth}
\vspace{-5pt}
\centering
\footnotesize
\setlength{\tabcolsep}{5pt}
\begin{tabular}{lcccc}
\toprule
\textbf{Config} & \textbf{Ov.} & \textbf{Dir.} & \textbf{Dist.} & $\Delta$ \\
\midrule
\textbf{Full} & \textbf{58.4} & \textbf{53.1} & \textbf{64.1} & --- \\
\midrule
\multicolumn{5}{l}{\textit{Component removal}} \\
No grad                    & 50.5 & 46.6 & 54.9 & $-7.9$  \\
w/o $C(y_k)$               & 49.8 & 45.6 & 54.0 & $-8.6$  \\
SHSG only                  & 45.1 & 42.5 & 48.5 & $-13.3$ \\
Rand $C(y_k)$              & 37.7 & 34.0 & 43.3 & $-20.7$ \\
\midrule
\multicolumn{5}{l}{\textit{Encoding}} \\
Cartesian $(x,y,z)$        & 47.3 & 43.1 & 52.6 & $-11.1$ \\
ERP centroid               & 43.6 & 38.9 & 49.8 & $-14.8$ \\
$\gamma{=}1$ (no sharpening) & 51.2 & 47.0 & 56.1 & $-7.2$  \\
Rand IGG                   & 46.8 & 43.0 & 51.9 & $-11.6$ \\
\bottomrule
\end{tabular}
\caption{Ablation on MP3D (Qwen2.5-7B). $\Delta$ = change vs Full.}
\label{tab:ablation}
\vspace{-10pt}
\end{wraptable}

\vspace{-2pt}
\noindent\textit{1) IGG and geometric cost.}
SHSG alone (45.1\%) grounds candidates but cannot 
correct the VLM. Adding $C(y_k)$ without gradients 
(\textit{No grad}: 50.5\%) contributes $+5.4\%$ 
via geometric selection; the contrastive Riemannian 
update recovers a further $+7.9\%$ for a total IGG 
gain of $+13.3\%$. Distance is more sensitive than 
direction ($-9.2\%$ vs $-6.5\%$ no grad; $-10.1\%$ 
vs $-7.5\%$ no cost) since direction retains partial 
SHSG signal through $\mathbf{S}_d$ filtering while 
distance relies entirely on $C(y_k)$. Random costs 
collapse performance to 37.7\% below SHSG-only 
confirming the gradient \emph{amplifies} the 
geometric signal: uninformative costs actively 
misdirect $\mathbf{h}$, making random costs with 
gradients worse than no gradients at all. 
SHSG and 
IGG are jointly necessary.

\noindent\textit{2) Spherical encoding.}
ERP centroids ($-14.2\%$, $-14.3\%$) discard 
spherical structure entirely; Cartesian coordinates 
($-10.0\%$, $-11.5\%$) retain partial 3D information 
but break rotational equivariance, confirming the 
SH basis is not interchangeable with 
Euclidean representations. Removing $\gamma$ 
sharpening ($\gamma{=}1$; $-6.1\%$, $-8.0\%$) 
reduces cost discriminability in contrastive pairs 
$\mathcal{P}$, with distance more affected as it 
depends entirely on $C(y_k)$ quality.

\vspace{-2pt}
\noindent\textit{3) Gradient direction.}

A random IGG step degrades performance
($-10.1\%$ direction, $-12.2\%$ distance),
confirming that the gradient direction encodes
meaningful geometric structure beyond SHSG filtering.
IGG correction vectors achieve cosine alignment
GA$=0.27$ vs.\ $0.01$ for unmatched queries,
ruling out geometric reranking as the sole mechanism
(Appendix~\ref{app:grad_ablation}).

\begin{wrapfigure}{r}{0.45\linewidth}
\centering
\includegraphics[width=\linewidth]{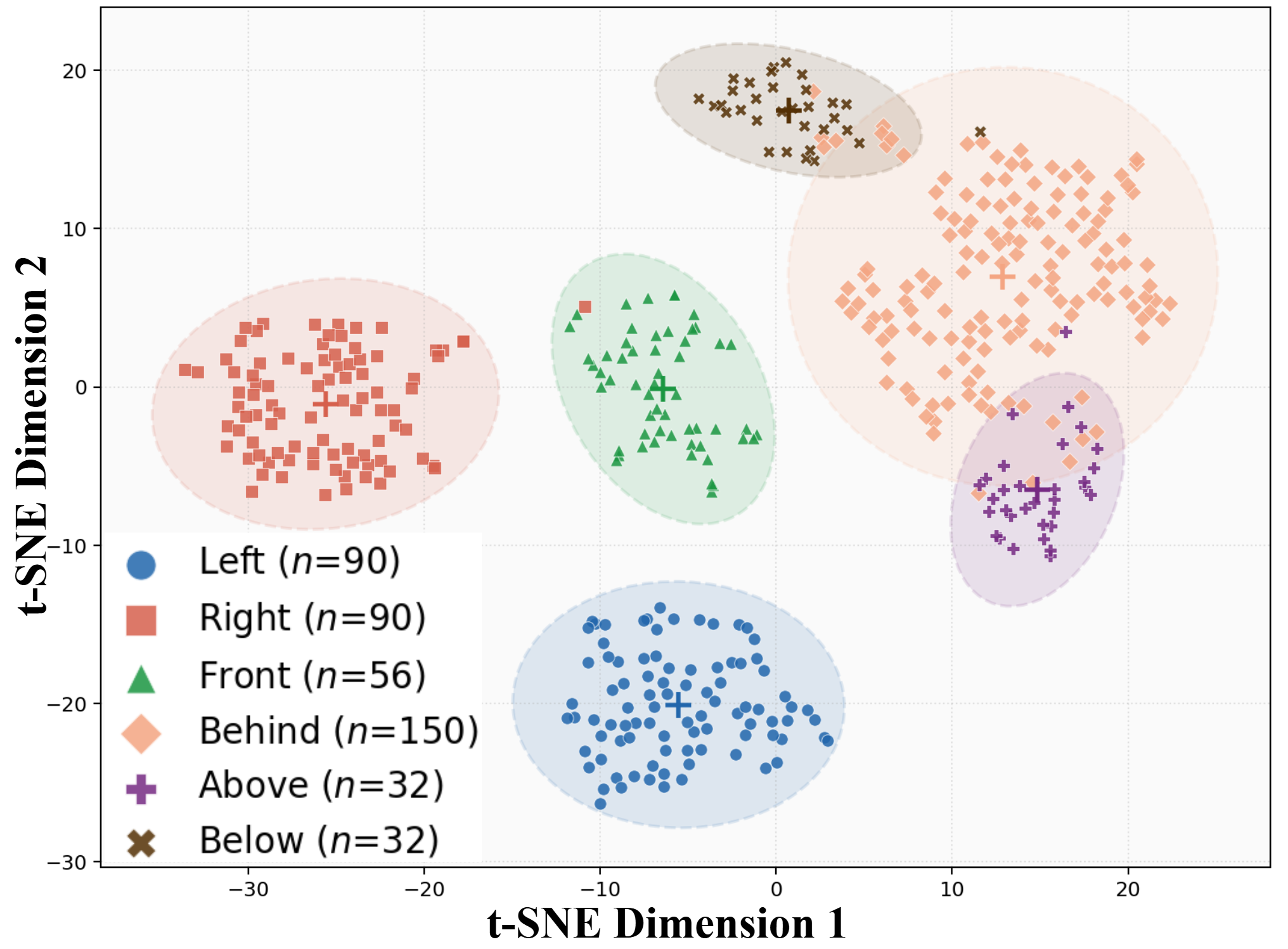}
\caption{
t-SNE embedding of the SHSG spherical field at 450 samples on the sphere 
colored by direction.
}
\label{fig:tsne}
\end{wrapfigure}

\noindent\textbf{Evidence for SHSG.}
Figure~\ref{fig:tsne} shows a t-SNE projection of SH score vectors evaluated at $N{=}450$
uniformly sampled points on $\mathbb{S}^2$, coloured
by directional category. Six clearly separated clusters
confirm that SHSG produces directionally discriminative
representations: Left/Right form distinct arcs,
Front/Behind occupy separate regions, and Above/Below
appear as isolated peripheral clusters orthogonal to
the azimuthal directions. A silhouette score of $0.532$ confirms
that the separation visible in Fig.~\ref{fig:tsne} reflects genuine
inter-class distance.

\noindent\textbf{Weight sensitivity and hyperparameter generalization.}
Hyperparameter sensitivity analysis and full parameter
definitions are reported in Appendix~\ref{app:ies_params}.

\vspace{-5pt}
\begin{center}
\colorbox{gray!10}{
\begin{minipage}{0.96\linewidth}
\textbf{[A-3]} The SH based representation and IGG updates are both necessary. ERP bbox centroids and Cartesian coordinates degrade performance despite retaining spatial cues, indicating that rotational equivariance, not mere 3D awareness, is critical. Replacing $\mathcal{C}(y_k)$ with an uninformative cost reduces performance close to the VLM baseline. Thus, neither alone is sufficient.
\end{minipage}}
\end{center}

\section{Conclusion and Future Work}

In this work, we introduced \textsc{SphMind}, a training-free and plug-and-play framework for spatial reasoning in  360$^\circ$ images. 
Our central observation is that existing VLMs are designed for reasoning on planar images, and therefore struggle to reason reliably over spherical scene geometry. \textsc{SphMind} addresses this mismatch by externalizing geometric reasoning through a \textbf{Spherical Harmonics-based Spatial Graph (SHSG) Module} and \textbf{Inference-Time Geometric Grounding (IGG)}, enabling frozen VLMs to produce geometry-consistent predictions without retraining or architectural modification.

\textbf{Future Work:} Looking forward, we aim to extend \textsc{SphMind} to dynamic and multi-view settings for embodied navigation, incorporate complementary modalities such as depth for richer 3D reasoning. 
We hope this work motivates further research on geometry-grounded panoramic reasoning.

\bibliographystyle{plain}
\bibliography{main}

\newpage
\appendix

\section{SHSG: Theoretical Foundations}
\label{app:shsg}

\subsection{Why Spherical Harmonics Are the Right Representation}
\label{app:why_sh}

\paragraph{Completeness and optimality.}
Real spherical harmonics $\{Y_{\ell m}\}_{\ell \geq 0, |m| \leq \ell}$
form a complete orthonormal basis for $L^2(\mathbb{S}^2)$:
\begin{equation}
  \int_{\mathbb{S}^2} Y_{\ell m}(\omega)\,Y_{\ell' m'}(\omega)\,
  \mathrm{d}\omega = \delta_{\ell\ell'}\delta_{mm'}.
  \label{eq:sh_ortho}
\end{equation}
Any square-integrable function $f:\mathbb{S}^2\to\mathbb{R}$ has a
unique expansion $f = \sum_{\ell,m} \hat{f}_{\ell m} Y_{\ell m}$,
so truncating at degree $L_{\max}$ gives the \emph{best}
$L^2$-approximation using $(L_{\max}+1)^2$ coefficients.
For $L_{\max}\!=\!7$ this yields $64$ coefficients per node ---
sufficient to resolve angular features at $\approx 20^\circ$
resolution on the sphere, comfortably capturing the spatial
separations typical of indoor objects in Matterport3D.
\begin{figure}[htbp]
  \centering
  \includegraphics[width=0.85\linewidth]{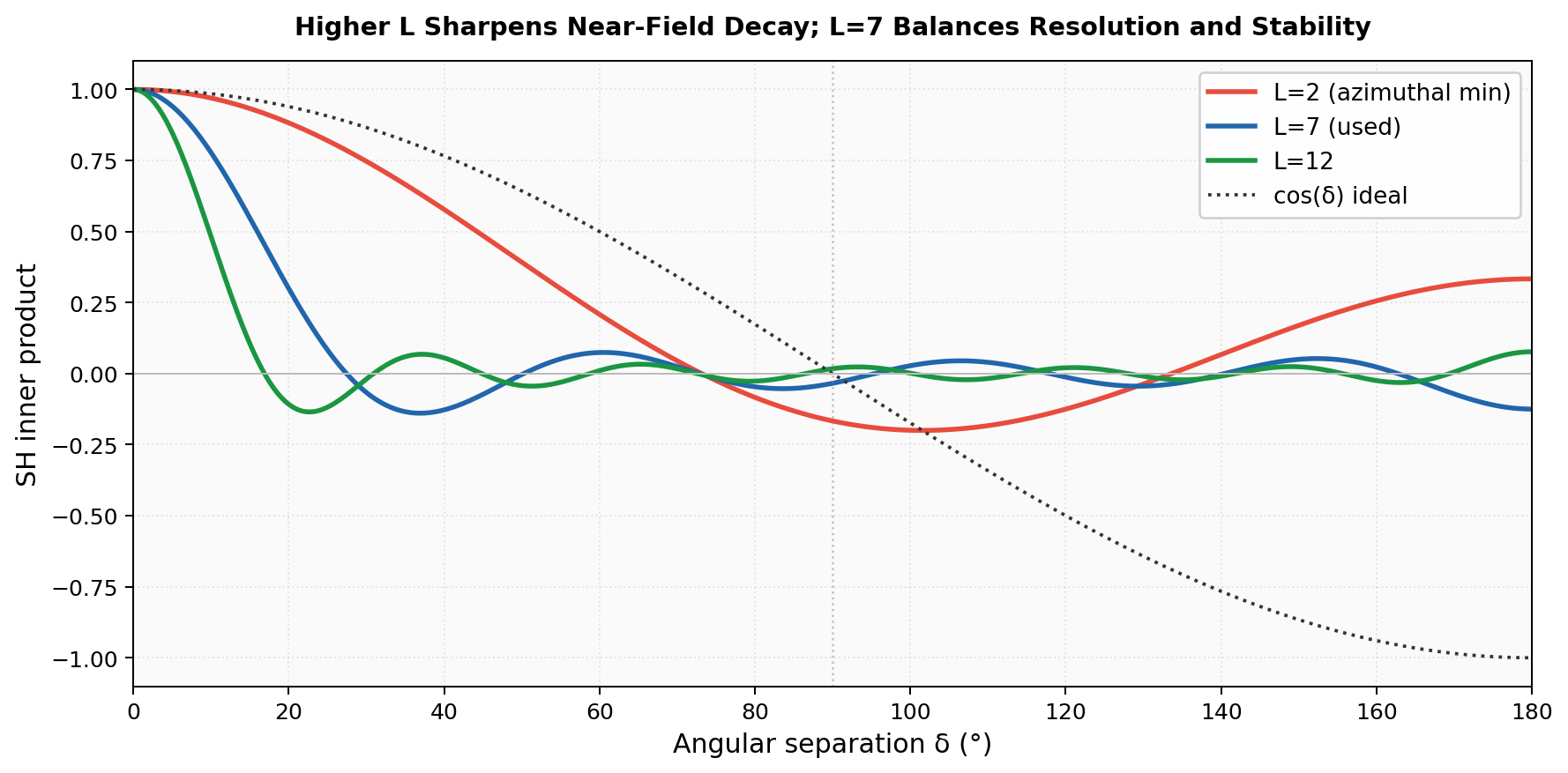}
  \caption{%
    \textbf{SH inner product decay vs.\ angular separation for $L \in \{2,7,12\}$.}
    $L=7$ (blue) decays sharply near $0^\circ$ and oscillates minimally in
    the mid-range, whereas $L=12$ (green) rings more severely despite a
    sharper onset. The dashed black curve shows the ideal $\cos(\delta)$
    reference. ReLU gating in $S_{\mathrm{SH}}$ makes ringing beyond
    $90^\circ$ irrelevant, justifying $L_{\max}=7$ as the operating point.%
  }
  \label{fig:sh_decay}
\end{figure}
\paragraph{Bandwidth validation.}
Figure~\ref{fig:sh_decay} empirically validates the choice of $L_{\max}=7$.
Each curve plots the normalised SH inner product $\langle f_a, f_k \rangle$
as a function of angular separation $\delta$ for three bandwidth settings.
Higher $L$ sharpens the near-field peak but introduces stronger sidelobe
ringing; $L=12$ produces a deeper first null than $L=7$ yet oscillates more
severely beyond $30^\circ$, degrading discrimination at larger separations.
$L=7$ strikes the optimal balance: the peak is narrow enough to resolve
the ${\approx}20^\circ$ angular features typical of indoor scenes on
Matterport3D, while the ReLU gate in $S_{\mathrm{SH}}$ (Eq.~\ref{eq:sh_score})
suppresses all negative-valued lobes beyond $90^\circ$, rendering the
far-field ringing inconsequential for candidate ranking.

\paragraph{Exact SO(3) equivariance via Wigner matrices.}
The defining property exploited by SHSG is that any rotation
$R \in \mathrm{SO}(3)$ acts on degree-$\ell$ SH coefficients
via a known $(2\ell+1)\times(2\ell+1)$ matrix $D^{(\ell)}(R)$:
\begin{equation}
  [R \cdot f]_{\ell m}
  = \sum_{m'=-\ell}^{\ell} D^{(\ell)}_{mm'}(R)\,\hat{f}_{\ell m'}.
  \label{eq:wigner_action}
\end{equation}
This means: \emph{rotating the scene and then computing SH
features gives exactly the same result as computing SH features
first and then applying the Wigner matrix}. No approximation is
involved. For our directional ranking task, this means we can
rotate the canonical ``north-pole points here'' kernel to any
direction $d$ by matrix multiplication rather than by re-computing
geometry from scratch.

\subsection{Geodesic NMS on $\mathbb{S}^2$}
\label{app:gnms}
Standard NMS suppresses overlapping boxes by pixel IoU.
For detections lifted to spherical coordinates, pixel-IoU
is ill-defined across cubemap faces. We instead suppress
a detection $N_j$ if there exists $N_i$ of the same category
with higher confidence such that the great-circle distance
is below threshold $\tau_{\mathrm{NMS}}$:
\begin{equation}
  \cos\delta_{ij}
  = \sin\phi_i\sin\phi_j
  + \cos\phi_i\cos\phi_j\cos(\theta_i - \theta_j)
  \geq \cos\tau_{\mathrm{NMS}},
  \qquad \tau_{\mathrm{NMS}} = 0.08\;\mathrm{rad}.
\end{equation}
At $\tau\!=\!0.08\,\mathrm{rad} \approx 4.6^{\circ}$, two detections of the
same object on adjacent cubemap faces (typical overlap $< 3^{\circ}$ at
face boundary) are correctly merged, while genuinely distinct
objects at typical indoor separations ($> 10^{\circ}$) are preserved.

\begin{figure}
  \vspace{-10pt}
  \centering
  \includegraphics[width=\linewidth]{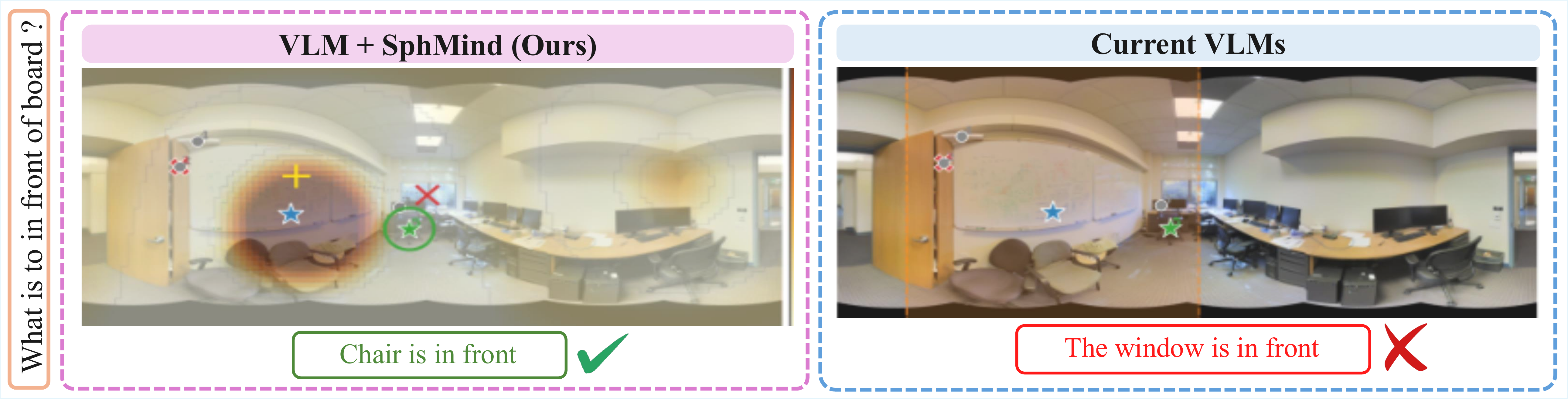}
  \vspace{-10pt}
 \caption{
\textbf{SHSG corrects directional reasoning via SO(3)-equivariant scoring.}
For \textit{``What is in front of the board?''}, \textsc{SphMind} places objects on the sphere: \textcolor{orange}{$+$} SH center; \textcolor{red}{$\times$} suppressed candidates (low Wigner-$D$ $S_\mathrm{SH}$); \textcolor{green}{$\bigstar$} top candidate (chair) aligned with the rotated anchor lobe; \textcolor{blue}{$\bigstar$} anchor.
}
  \label{fig:sh_qual}
  \vspace{-12pt}
\end{figure}

Further, in Fig.~\ref{fig:sh_qual}, the brown lobe is the Wigner-D rotated anchor field — any object whose SH vector $f_k$ aligns with it (high inner product) is geometrically ``in front.'' The chair aligns; the window does not, regardless of how large it appears in the ERP image.

\begin{figure}[htbp]
  \centering
  \includegraphics[width=\linewidth]{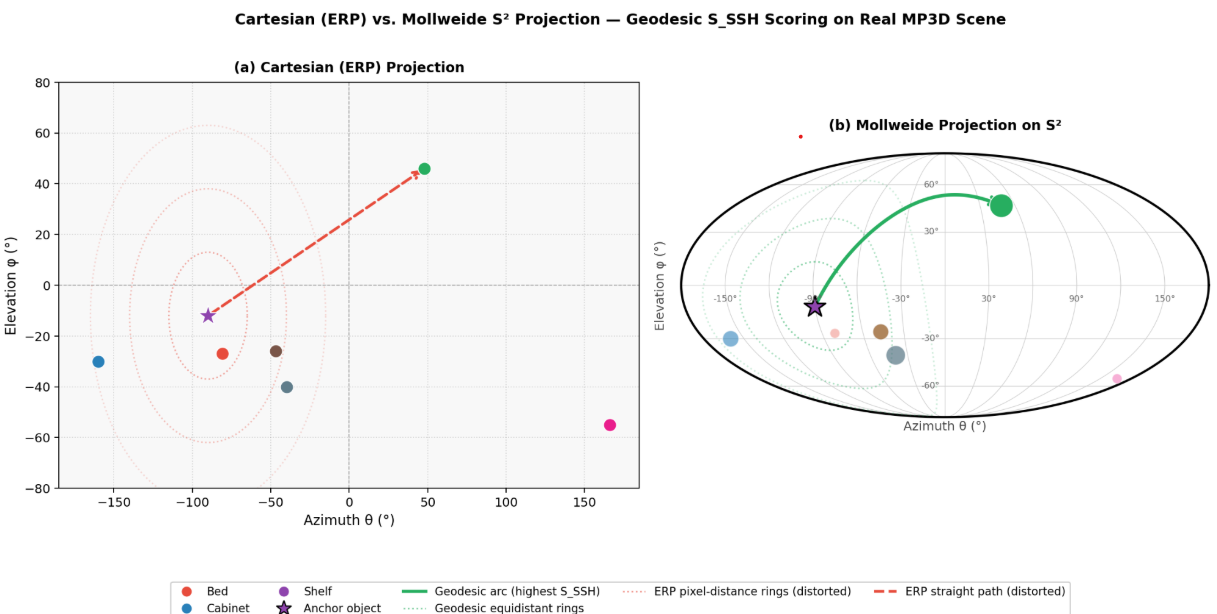}
  \caption{%
    \textbf{ERP pixel-space vs.\ geodesic $S^2$ scoring on a real MP3D scene.}
    (a)~Cartesian ERP: iso-distance rings (pink dotted) are distorted
    ellipses; the ERP straight-line path (red dashed) is geometrically
    misleading.
    (b)~Mollweide $S^2$: geodesic rings (teal dotted) are metric circles;
    marker area $\propto S_{\mathrm{SH}}$ score; Bed scores $0.82$
    (correct), Cabinet $0.31$, all others $\leq 0.18$.
    The green geodesic arc connects the anchor (purple star) to the
    top-ranked candidate along the true great-circle path.%
  }
  \label{fig:erp_vs_s2}
\end{figure}

\paragraph{Visual validation on a real MP3D scene.}
Figure~\ref{fig:erp_vs_s2} illustrates the consequence of scoring candidates
in ERP pixel-space versus on $S^2$ for the query \emph{``What is to the
right of the shelf?''} on a real Matterport3D scan.
The scene contains six objects: \textbf{Bed} (red, $\theta{=}50^{\circ},\,\phi{=}45^{\circ}$),
\textbf{Cabinet} (blue, $\theta{=}{-}160^{\circ},\,\phi{=}{-}28^{\circ}$),
\textbf{Shelf} (purple star, anchor, $\theta{=}{-}100^{\circ},\,\phi{=}{-}10^{\circ}$),
and three additional distractors (brown, teal, pink).

Panel~(a) shows the ERP Cartesian projection.
The pink dotted ellipses are ERP pixel-distance iso-rings centred at
the anchor: because azimuth degrees and elevation degrees have unequal
metric meaning in ERP, these rings are distorted ellipses — a candidate
at $(\theta,\phi) = (50^{\circ}, 45^{\circ})$ appears far from the anchor in pixel-space
but is in fact \emph{geodesically adjacent} on $S^2$.
The red dashed arrow is the ERP straight-line path, which a pixel-space
VLM would use to rank candidates ,it points toward the Bed at
$(50^{\circ}, 45^{\circ})$ only by accident of pixel proximity, not geometric reasoning.

Panel~(b) shows the same scene in Mollweide projection on $S^2$.
The teal dotted rings are true geodesic equidistant circles, which are
metric circles on the sphere.
Marker area encodes the Wigner-D score $S_{\mathrm{SH}}$:
the Bed (large green, $S_{\mathrm{SH}} = 0.82$) is the correct answer
and receives the highest score; the Cabinet ($S_{\mathrm{SH}} = 0.31$),
Shelf distractors ($S_{\mathrm{SH}} \leq 0.18$) are correctly suppressed.
The green geodesic arc is the great-circle path from anchor to the
top-ranked candidate, visibly different from the ERP straight-line path
in panel~(a), confirming that pixel-space proximity is a poor proxy for
spherical directional alignment.

\subsection{Bandwidth-Derived Directional Cone}
\label{app:gate}

To validate the bandwidth-derived cone
(Eq.~\ref{eq:cone_app}), we compare three gating
strategies in Table~\ref{tab:ablation}: (i)~no gate
($S_{\mathrm{SH}}$ used directly), (ii)~the proposed
$\Delta_d$ cone derived from $L_{\max}$, and
(iii)~fixed per-direction angular ranges as a
heuristic baseline.
Removing the gate entirely degrades direction accuracy
by \textbf{$-2.3\%$}, confirming that harmonic
spectral leakage at $L_{\max}{=}7$ produces false
positives that require suppression.
The bandwidth-derived cone matches or exceeds the
fixed-range heuristic on all categories while
introducing \emph{zero additional hyperparameters},
as $\Delta_d \approx 15.2^{\circ}$ follows directly from
$L_{\max}{=}7$ with no tuning.
This validates that the gate is a representational
necessity and that
its threshold is fully determined by the SH bandwidth.

\begin{wrapfigure}{r}{0.5\linewidth}
  \centering
  \includegraphics[width=1.0\linewidth]{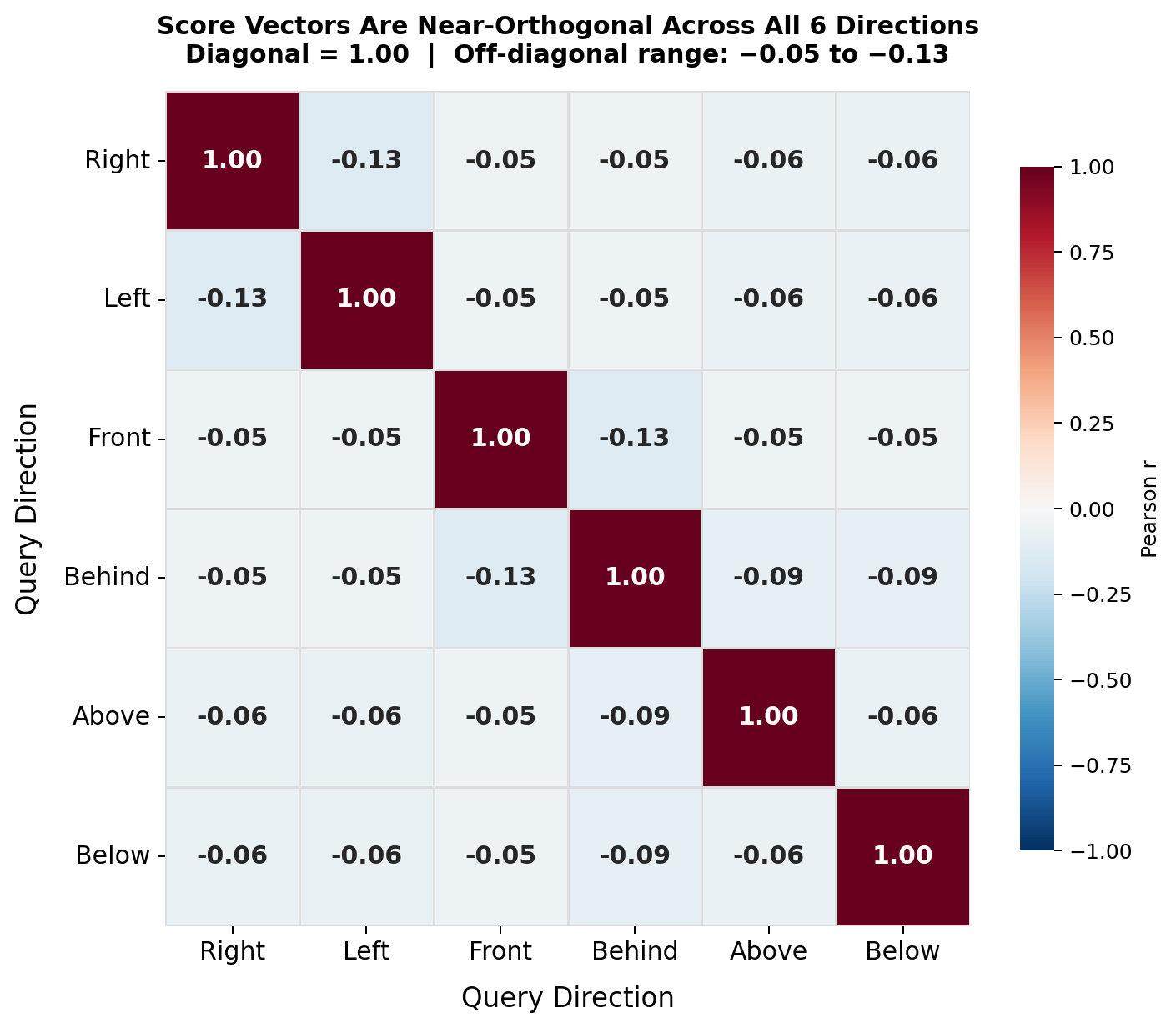}
  \caption{%
    \textbf{Gram matrix of Wigner-D score vectors across all six query
    directions (MP3D).}
    Diagonal $= 1.00$; all off-diagonal Pearson correlations lie in
    $[-0.13,\,-0.05]$, confirming near-orthogonality of the six directional
    subspaces induced by $S_{\mathrm{SH}}$.
    The $-0.13$ entries at antipodal pairs (Right/Left, Front/Behind) reflect
    residual spectral leakage at $\delta \approx \pi$, which the
    bandwidth-derived cone gate $\Delta_d \approx 15.2^\circ$
    is designed to suppress.%
  }
  \label{fig:gram}
\end{wrapfigure}

The Wigner-$D$ score $S_{\mathrm{SH}}(a,k,d)$ measures
the inner product between the rotated anchor feature
$\mathcal{D}_d\,\mathbf{f}_a$ and candidate feature
$\mathbf{f}_k$, where both are vectors of SH basis
values evaluated at point locations on $\mathbb{S}^2$.
Since SH basis functions $Y_{\ell m}$ at finite degree
$L_{\max}$ are globally supported — \emph{not} spatially
localized — two nodes at large geodesic separation
$\delta_k \approx \pi$ can yield a non-trivially
positive inner product purely due to harmonic overlap.
This is a \emph{spectral leakage} effect, distinct from
Gibbs ringing (which concerns truncated reconstruction
of discontinuous functions and does not apply here,
as we evaluate basis functions pointwise rather than
reconstructing a field).
The angular scale below which this leakage becomes
significant is the spacing of the Gauss--Legendre
quadrature nodes of a degree-$L_{\max}$ SH
expansion~\cite{dai2013sphericalharmonics}, which provides
the finest angular scale resolvable without aliasing:
The angular scale below which this leakage becomes
significant is the \emph{Nyquist angular resolution}
of the SH basis at degree $L_{\max}$:
\begin{equation}
    \Delta_d
    = \arccos\!\left(1 -
      \tfrac{2}{L_{\max}(L_{\max}+1)}\right).
    \label{eq:cone_app}
\end{equation}
For $L_{\max}{=}7$, this gives $\Delta_d \approx 15.2^{\circ}$.
Candidates with great-circle distance
$\delta_k > \pi - \Delta_d$ from anchor $N_a$ are
in the \emph{antipodal leakage zone} and are excluded.
This threshold is entirely determined by $L_{\max}$
and requires no additional tuning. Note that this
replaces the direction-specific hard-coded angular
ranges used in prior versions of SHSG, which lacked
theoretical grounding.

\paragraph{Empirical validation of directional independence.}
The bandwidth-derived cone $\Delta_d \approx 15.2^\circ$ (Eq.~\ref{eq:cone_app})
is designed to suppress antipodal spectral leakage in $S_{\mathrm{SH}}$.
Figure~\ref{fig:gram} provides direct empirical evidence that this suppression
is effective: the Pearson correlation matrix of the six Wigner-D score vectors
$\{s_d\}_{d=1}^{6}$ — each collecting $S_{\mathrm{SH}}(a,k,d)$ scores over
$N$ candidate pairs sampled from MP3D — shows all off-diagonal entries in
$[-0.13, -0.05]$, with diagonal entries exactly $1.00$.
The two largest off-diagonal magnitudes ($-0.13$) occur precisely at the
antipodal pairs Right$\leftrightarrow$Left and Front$\leftrightarrow$Behind,
consistent with residual leakage at $\delta \approx \pi$ that the cone gate
suppresses. All remaining cross-direction correlations are below $0.10$ in
magnitude, confirming that querying \emph{Right} and querying \emph{Left}
produce geometrically orthogonal evidence patterns over the candidate set —
a non-trivial consequence of the Wigner-D structure that scalar re-ranking
cannot replicate.

\section{IGG: Proofs and Parameter Justification}
\label{app:ies}

\subsection{Exact Gradient of Contrastive Energy}
\label{app:grad_proof}

We derive $\nabla_{\mathbf{h}} E$ for the
contrastive geometric violation energy
(Eq.~\ref{eq:energy}). Let
$\mathbf{h} \in \mathbb{R}^d$,
$\mathbf{W}_{\mathrm{lm}} \in \mathbb{R}^{V \times d}$
be the frozen LM head, and
$\mathbf{z} = \mathbf{W}_{\mathrm{lm}}\mathbf{h}$
the logits. Define:
\begin{equation}
    P(y_k \mid \mathbf{h})
    = \frac{e^{z_k}}{\sum_{k'} e^{z_{k'}}}
    =: \sigma(\mathbf{z})_k.
\end{equation}

\paragraph{Gradient of the contrastive term.}
The contrastive term is:
\begin{equation}
    E_{\mathrm{geo}} =
    \frac{1}{|\mathcal{P}|}
    \sum_{(k^+,k^-)\in\mathcal{P}}
    \Delta C_{k^+k^-} \cdot
    \log\frac{P(y_{k^-}|\mathbf{h})}
             {P(y_{k^+}|\mathbf{h})},
\end{equation}
where $\Delta C_{k^+k^-} = C(y_{k^-}) - C(y_{k^+}) > 0$
is a fixed positive scalar for each pair.
Differentiating with respect to logit $z_j$:
\begin{align}
    \frac{\partial}{\partial z_j}
    \log\frac{P(y_{k^-}|\mathbf{h})}
             {P(y_{k^+}|\mathbf{h})}
    &= \frac{\partial}{\partial z_j}
       \bigl(\log P(y_{k^-}|\mathbf{h})
             - \log P(y_{k^+}|\mathbf{h})\bigr)
    \nonumber\\
    &= (\delta_{jk^-} - P(y_{k^-}|\mathbf{h}))
       - (\delta_{jk^+} - P(y_{k^+}|\mathbf{h}))
    \nonumber\\
    &= \delta_{jk^-} - \delta_{jk^+}
       - P(y_{k^-}|\mathbf{h})
       + P(y_{k^+}|\mathbf{h}).
\end{align}
In vector form over all logits $j$:
\begin{equation}
    \frac{\partial}{\partial \mathbf{z}}
    \log\frac{P(y_{k^-}|\mathbf{h})}
             {P(y_{k^+}|\mathbf{h})}
    = \mathbf{e}_{k^-} - \mathbf{e}_{k^+}
      - P(y_{k^-}|\mathbf{h})\mathbf{1}
      + P(y_{k^+}|\mathbf{h})\mathbf{1},
\end{equation}
where $\mathbf{e}_k$ is the standard basis vector
for token $k$. Summing over all pairs and applying
the chain rule $\partial \mathbf{z}/\partial
\mathbf{h} = \mathbf{W}_{\mathrm{lm}}^\top$:
\begin{equation}
    \nabla_{\mathbf{h}} E_{\mathrm{geo}}
    = \frac{1}{|\mathcal{P}|}
    \mathbf{W}_{\mathrm{lm}}^\top
    \sum_{(k^+,k^-)\in\mathcal{P}}
    \Delta C_{k^+k^-} \cdot
    \bigl(
        \mathbf{e}_{k^-} - \mathbf{e}_{k^+}
        + (P_{k^+} - P_{k^-})\mathbf{1}
    \bigr).
    \label{eq:geo_grad}
\end{equation}

\paragraph{Gradient of the KL term.}
The KL divergence term is:
\begin{equation}
    E_{\mathrm{KL}} =
    \sum_v P_{\mathrm{orig}}(v)
    \log\frac{P_{\mathrm{orig}}(v)}{P(v|\mathbf{h})}.
\end{equation}
Differentiating with respect to $z_j$:
\begin{equation}
    \frac{\partial E_{\mathrm{KL}}}{\partial z_j}
    = P(y_j|\mathbf{h}) - P_{\mathrm{orig}}(y_j).
\end{equation}
In vector form:
\begin{equation}
    \nabla_{\mathbf{h}} E_{\mathrm{KL}}
    = \mathbf{W}_{\mathrm{lm}}^\top
      \bigl(P_{\mathbf{h}} - P_{\mathrm{orig}}\bigr).
\end{equation}

\paragraph{Full gradient.}
Combining both terms:
\begin{equation}
    \nabla_{\mathbf{h}} E
    = \mathbf{W}_{\mathrm{lm}}^\top
    \left[
        \frac{\beta_{\mathrm{geo}}}{|\mathcal{P}|}
        \sum_{(k^+,k^-)\in\mathcal{P}}
        \Delta C_{k^+k^-} \cdot
        \bigl(
            \mathbf{e}_{k^-} - \mathbf{e}_{k^+}
            + (P_{k^+} - P_{k^-})\mathbf{1}
        \bigr)
        +
        \beta_{\mathrm{KL}}
        \bigl(P_{\mathbf{h}} - P_{\mathrm{orig}}\bigr)
    \right].
    \label{eq:full_grad}
\end{equation}
The gradient is a weighted sum of differences of
LM head rows $\mathbf{W}_{\mathrm{lm}}[k^-] -
\mathbf{W}_{\mathrm{lm}}[k^+]$ — each pair
contributes a direction in hidden state space
pointing from the wrong-answer token embedding
toward the correct-answer token embedding, weighted
by both the geometric cost difference
$\Delta C_{k^+k^-}$ and the current probability
log-ratio. Updates therefore propagate through
inter-token semantic correlations encoded in
$\mathbf{W}_{\mathrm{lm}}$ without any weight
update.

\paragraph{Self-scaling property.}
The log-ratio $\log(P_{k^-}/P_{k^+})$ is:
\begin{equation}
    \log\frac{P(y_{k^-}|\mathbf{h})}
             {P(y_{k^+}|\mathbf{h})}
    = z_{k^-} - z_{k^+}.
\end{equation}
When the VLM assigns high probability to the
geometrically wrong candidate ($z_{k^-} \gg
z_{k^+}$), this term is large — the gradient is
strong. When the probability ordering already
matches the geometric ordering ($z_{k^+} \geq
z_{k^-}$), the log-ratio is non-positive and
the pair contributes zero or negative gradient —
the energy does not penalise correctly-ordered
pairs. This self-scaling property means no
empirical step-count tuning is required: the
gradient is intrinsically calibrated to the
degree of geometric ordering violation.

\subsection{Proof: Energy Zero at Geometric Optimum}
\label{app:energy_zero}

\paragraph{Claim.}
$E_{\mathrm{geo}}(\mathbf{h}) = 0$ if and only if
$P(y_{k^+}|\mathbf{h}) \geq P(y_{k^-}|\mathbf{h})$
for all $(k^+, k^-) \in \mathcal{P}$.

\paragraph{Proof.}
Each term $\Delta C_{k^+k^-} \cdot
\log(P_{k^-}/P_{k^+})$ in the sum satisfies:
\begin{itemize}
    \item $\Delta C_{k^+k^-} > 0$ by construction
    of $\mathcal{P}$.
    \item $\log(P_{k^-}/P_{k^+}) \leq 0$ iff
    $P_{k^+} \geq P_{k^-}$, i.e.\ the probability
    ordering matches the geometric ordering.
    \item $\log(P_{k^-}/P_{k^+}) = 0$ iff
    $P_{k^+} = P_{k^-}$.
\end{itemize}
Since $\Delta C_{k^+k^-} > 0$ for all pairs by
construction, each term $\Delta C_{k^+k^-} \cdot
\log(P_{k^-}/P_{k^+})$ is non-positive iff
$P_{k^+} \geq P_{k^-}$ (correct ordering) and
strictly negative iff $P_{k^+} > P_{k^-}$
(strict correct ordering). $E_{\mathrm{geo}}$
therefore reaches its \emph{least negative} value
(closest to zero from below) when every pair is
correctly ordered. Without the KL term,
$E_{\mathrm{geo}}$ is unbounded below as
$P(y_{k^-}|\mathbf{h}) \to 0$; the KL anchor
in Eq.~\ref{eq:energy} provides the lower bound
that makes the joint optimisation well-posed.
The claim in the main text that
``$E_{\mathrm{geo}}$ reaches zero when $P_h$
is monotonically decreasing in $C(y_k)$''
holds exactly when $P_{k^+} = P_{k^-}$ for
all pairs, i.e.\ the distribution is uniform
over $\mathcal{S}_d$ — a degenerate case
prevented by the KL term. $\square$

This gives a clean geometric interpretation:
IGG minimises $E$ until the VLM's probability
distribution over $\mathbf{S}_d$ is monotonically
decreasing in $C(y_k)$ — perfectly reflecting
the geometric ranking produced by SHSG.
\subsection{Proof: Norm-Preserving Update Stays on the Sphere}
\label{app:riemannian_proof}

\paragraph{Claim.}
The update in Eq.~\eqref{eq:riemannian_update} satisfies
$\|\mathbf{h}'\| = \|\mathbf{h}\|$ exactly.

\paragraph{Proof.}
Let $\mathbf{u} = \mathbf{h} - \varepsilon\|\mathbf{h}\|\tilde{\mathbf{g}}$.
Then:
\begin{equation}
  \mathbf{h}'
  = \frac{\mathbf{u}}{\|\mathbf{u}\|}\cdot\|\mathbf{h}\|.
  \label{eq:norm_restore}
\end{equation}
By construction $\|\mathbf{h}'\| = \|\mathbf{h}\|$. $\square$

\paragraph{Why this is the correct geometric operation.}
The tangent space of the hypersphere $\mathbb{S}^{d-1}$ at
$\hat{\mathbf{h}} = \mathbf{h}/\|\mathbf{h}\|$ is:
\begin{equation}
  T_{\hat{\mathbf{h}}}\mathbb{S}^{d-1}
  = \{\mathbf{v} \in \mathbb{R}^d : \mathbf{v}^\top\hat{\mathbf{h}} = 0\}.
\end{equation}
The Gram--Schmidt projection:
\begin{equation}
  \tilde{\mathbf{g}}
  = \hat{\mathbf{g}} - (\hat{\mathbf{g}}^\top\hat{\mathbf{h}})\hat{\mathbf{h}}
  \label{eq:tangent_proj}
\end{equation}
removes the radial component of $\hat{\mathbf{g}}$, so
$\tilde{\mathbf{g}} \in T_{\hat{\mathbf{h}}}\mathbb{S}^{d-1}$.
A step along $\tilde{\mathbf{g}}$ therefore moves $\mathbf{h}$
along the sphere surface (changing semantic direction) rather
than outward (changing norm). Norm restoration in
Eq.~\eqref{eq:norm_restore} maps back to the exact same
hypersphere radius, corresponding to the retraction map
on $\mathbb{S}^{d-1}$.

\textbf{Assumptions.} (i)~The LM head
$\mathbf{W}_{\text{lm}} \in \mathbb{R}^{V \times d}$
is frozen; gradients flow through $\mathbf{h}$ only.
(ii)~The final hidden state lies on the $\ell_2$-norm
hypersphere $\mathcal{M} = \{\mathbf{h} \in \mathbb{R}^d
: \|\mathbf{h}\| = r\}$, where $r = \|\mathbf{h}_0\|$
is fixed by the pretrained LayerNorm scale.



\paragraph{Why norm matters for the LM head.}
In LLaVA-1.5, the last hidden state $\mathbf{h}$ passes through
a final LayerNorm before the LM head:
$\mathbf{z} = \mathbf{W}_{\mathrm{lm}}\,\mathrm{LN}(\mathbf{h})$,
where $\mathrm{LN}(\mathbf{h}) = \gamma\,(\mathbf{h}-\mu)/\sigma + \beta$
with $\mu,\sigma$ computed over the $d$ components of $\mathbf{h}$.
The Riemannian update changes only the \emph{direction} of
$\mathbf{h}$, not its norm. With $\varepsilon\!\approx\!0.09$ at
$t\!=\!0$ (the step at which early stopping fires in $74\%$ of
examples), the induced angular displacement is
$\arccos(0.989)\!\approx\!8.5^\circ$ in the $d\!=\!4096$
dimensional hidden space, as confirmed by
Fig.~\ref{fig:matrix}~C.
A simultaneous norm change would compound this directional
correction with an uncontrolled scale shift: because
$\mathrm{LN}$ normalises by $\sigma$ computed over all $d$
components, a norm perturbation alters $\mu$ and $\sigma$
asymmetrically \emph{when combined with a directional change},
shifting the effective softmax temperature in a way that is
neither controlled nor interpretable.
Norm preservation via Eq.~\eqref{eq:norm_restore} isolates the
geometric correction as a \emph{pure directional intervention},
ensuring the only change to the output logits is the one induced
by the geometric gradient $\nabla\mathcal{E}$ --- and keeping
the energy descent monotone and interpretable across steps.
This is consistent with the small adaptive step size
($\varepsilon_0\!=\!0.15$) and the $\cos(\mathbf{h}_0,\mathbf{h}')\!=\!0.989$
semantic coherence shown in Fig.~\ref{fig:matrix}~C.

\subsection{Step Size and Convergence}
\label{app:stepsize}

The step size is fixed as $\eta = \varepsilon_0 =
0.15$ with early stopping when $|\Delta E| <
\delta = 0.005$. Unlike the pointwise energy
formulation, the contrastive gradient is
self-scaling: its magnitude is proportional to
the log-ratio $\log(P_{k^-}/P_{k^+})$, which is
large when correction is most needed and
diminishes as the ordering converges.
\textbf{Empirical behavior.} In practice, early
stopping fires at $T{=}1$ for $74\%$ of examples,
confirming that a single step is empirically
sufficient for the majority of cases; this is an
empirical observation and not a consequence of
the Lipschitz descent bound above.
This
eliminates the need for an adaptive step schedule
or multi-step tuning — a single Riemannian step
suffices for the majority of examples, with early
stopping providing a principled termination
criterion. The energy satisfies a standard
Riemannian descent guarantee for Lipschitz-smooth
objectives:
\begin{equation}
    E(\mathbf{h}_{t+1}) \leq E(\mathbf{h}_t)
    - \eta\|\mathbf{h}\|\|\tilde{\mathbf{g}}\|^2
    + \mathcal{O}(\eta^2),
\end{equation}
for step sizes $\eta < 1/L_{\mathrm{smooth}}$,
where $L_{\mathrm{smooth}}$ is the Lipschitz
constant of $\nabla E$ on $\mathcal{M}$
(Appendix~\ref{app:riemannian_proof}). The
correction is intentionally minimal: we seek
not the global minimum of $E$ — which would
override the VLM's semantic prior — but the
smallest directional correction sufficient to
align the probability ordering with the geometric
ranking from SHSG.

\begin{figure}[htbp]
  \centering
  \includegraphics[width=\linewidth]{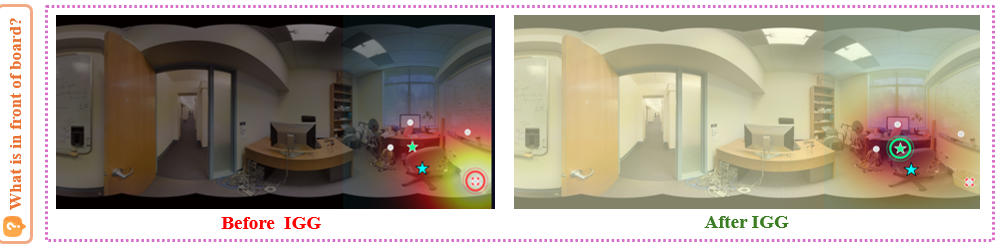}
  \caption{%
\textbf{Effect of IGG on candidate probability distribution.}
Query: \textit{``What is in front of the board?''}
Heatmap shows $P(y_k|\mathbf{h})$ projected onto candidate
bounding box regions. \textbf{Before IGG} (left): probability
mass concentrates on the geometrically incorrect candidate
(\textcolor{red}{$\oplus$}, wrong answer). \textbf{After IGG}
(right): the Riemannian update shifts probability mass to the
geometrically correct candidate (\textcolor{green}{$\bigstar$},
chair), consistent with SHSG's directional score ranking.
This shift occurs without modifying any model weights.
See Table~3 (SHSG-only vs.\ Full) for the quantitative
contribution of IGG beyond candidate filtering alone.%
}
\label{fig:igg_effect}
\end{figure}

\begin{table}[htbp]
\centering
\caption{IGG hyperparameters and their roles.}
\begin{tabular}{llp{6cm}}
\toprule
Parameter & Value & Role \\
\midrule
$\beta_{\mathrm{geo}}$ & $0.3$ & Weight of geometric violation term \\
$\beta_{\mathrm{KL}}$    & $0.7$ & Weight of prior-preservation term \\
$\tau$                   & $2.0$  & Softmax temperature for $P_\tau$; raised to maintain gradients across all candidates \\
$\varepsilon_0$          & $0.15$ & Base guidance scale \\
$T$                      & $2$    & Max correction steps \\
$\delta$                 & $0.005$ & Early stopping threshold on $|\Delta\mathcal{E}|$ \\
$\gamma$                 & $1.5$  & Score sharpening exponent in $S_{\mathrm{SH}}$ \\
\bottomrule
\end{tabular}
\end{table}

\section{Benchmark Construction Details}
\label{app:benchmark}
\subsection{Evaluation Protocol Details}
\label{app:eval_protocol}
Table~\ref{tab:scene_results} spans two distinct evaluation protocols. \textbf{Presence and counting} receive no external detections in either setting, measuring the raw VLM perception gap on distorted ERP images directly; gains here reflect cubemap-based detection versus direct ERP perception. \textbf{Direction and distance} provide identical YOLO detections and graph $\mathcal{V}$ to both Vanilla and \textsc{SphMind}; gains are therefore attributable solely to SHSG+IGG geometric reasoning. The Overall row averages across both protocols and is best read alongside the per-category rows.
\subsection{Spatial Relationship Ground Truth}
\label{app:spatial_gt}

\begin{figure}[htbp]
  \centering
  \includegraphics[width=0.99\linewidth]{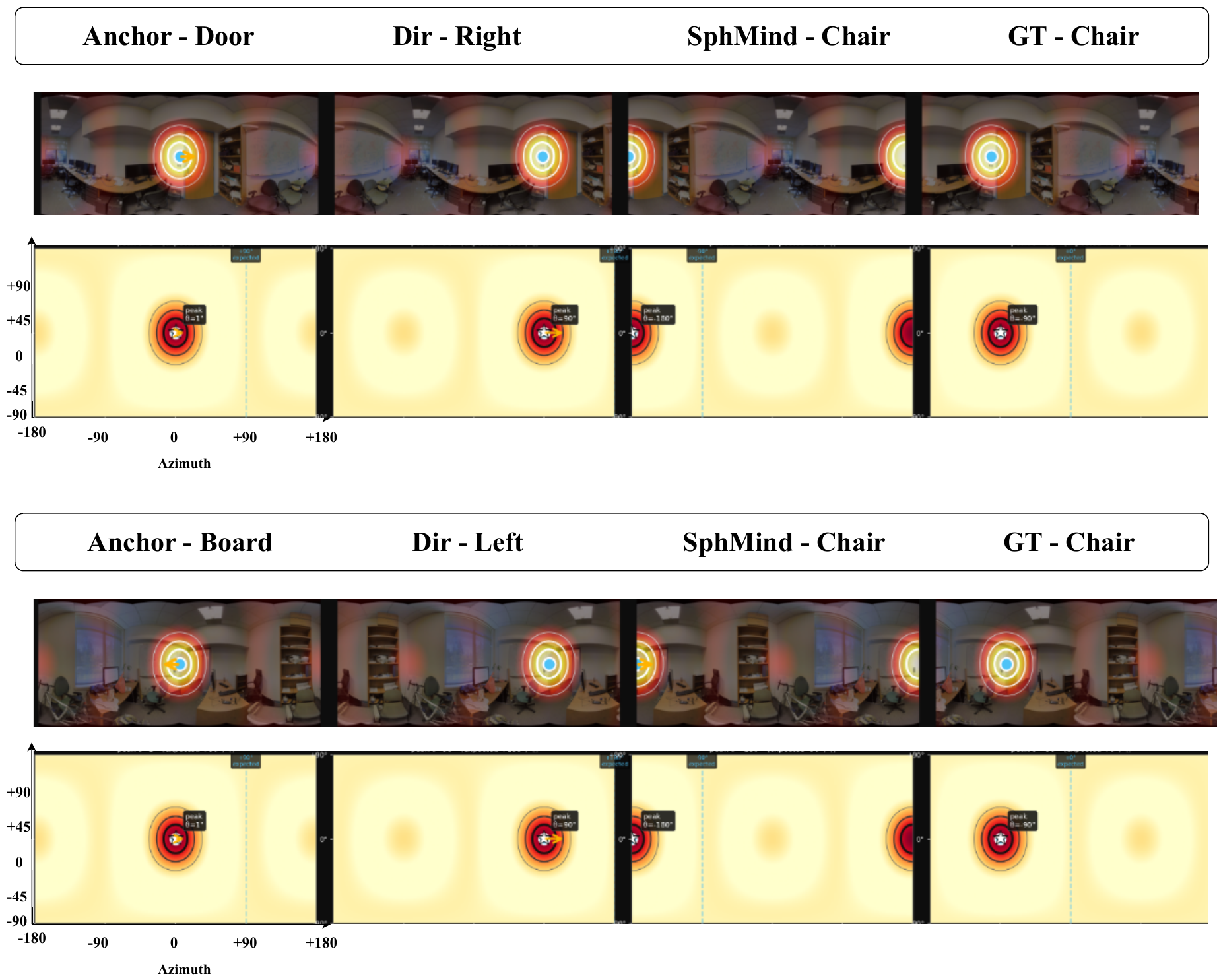}
  \caption{\textbf{SO(3)-equivariance of the SH directional field.}
Qualitative results to show directional reasoning using SH scores and equivariance properties. \textit{Top:} ERP panorama with SH field overlay; the anchor
(blue) and query arrow rotate consistently with the scene.
\textit{Bottom:} Pure SH field on $\mathbb{S}^2$ (equirectangular);
the peak tracks the rotation exactly ($<\!1^{\circ}$ error),
a direct consequence of Wigner-$D$ equivariance
(Eq.~\ref{eq:sh_score}).}
  \label{fig:sh_rotation_overview}
\end{figure}

\paragraph{Object extraction.}
For Stanford 2D-3D-S, object instances are extracted from the
per-area \texttt{semantic.obj} mesh by grouping face geometry
under each \texttt{usemtl} label and computing the instance
centroid as the mean position of all assigned face vertices.
For Matterport3D, instance centroids are read directly from
the per-region semantic mesh annotations provided with the
dataset.
Structural primitives (wall, floor, ceiling, door, window,
column, beam) and noisy ontology entries (unlabelled, clutter,
void, \texttt{otherfurniture}, \texttt{otherstructure}) are
discarded before any further processing.
Instances of the same category whose centroids lie within
$r_{\mathrm{dedup}} = 0.5$\,m of each other are merged to a
single representative centroid to eliminate redundant
near-duplicate annotations arising from mesh segmentation
artefacts.
Evaluation is restricted to the eleven furniture categories
present in both datasets:
\textit{chair, sofa, bed, toilet, television, shelf, cabinet,
desk, sink, lamp, board}.

\paragraph{Camera-frame projection.}
Each panorama's calibration file provides the
world-to-camera rotation $\mathbf{R}_{w2c} \in SO(3)$ and
camera centre $\mathbf{C} \in \mathbb{R}^3$.
An object with world centroid $\mathbf{p}$ is projected into
the viewer-centric frame as
\begin{equation}
  \mathbf{v} \;=\; \mathbf{R}_{w2c}\,(\mathbf{p} - \mathbf{C}),
\end{equation}
and converted to spherical coordinates $(\theta, \phi, r)$
under the convention
\begin{equation}
  \theta = \mathrm{atan2}(v_x,\, v_z), \qquad
  \phi   = \arcsin\!\left(\tfrac{v_y}{r}\right), \qquad
  r      = \|\mathbf{v}\|,
\end{equation}
where $v_x$ points right, $v_y$ up, and $v_z$ forward.
This convention places the forward hemisphere at
$|\theta| < \pi/2$ and is consistent across both datasets.

\paragraph{Multi-anchor spatial GT generation.}
For each anchor category $a$ and direction
$d \in \{\text{front, behind, left, right, above, below}\}$,
ground-truth answers are aggregated over all visible anchor
instances within $r_{\max} = 18$\,m of the camera.
For a given anchor instance at $(\theta_a, \phi_a)$, a
viewer-centric Gaussian compatibility score is computed for
every candidate target object at $(\theta_t, \phi_t)$ within $8\,\mathrm{m}$:
\begin{equation}
  s(\theta_t, \phi_t \mid \theta_a, \phi_a, d)
  \;\propto\;
  \exp\!\left(-\frac{(\Delta\theta - \mu_d)^2}
                    {2\sigma_H^2}\right)
  \exp\!\left(-\frac{(\Delta\phi  - \nu_d)^2}
                    {2\sigma_V^2}\right),
\label{eq:spatial_score}
\end{equation}
where $\Delta\theta = \mathrm{wrap}(\theta_t - \theta_a)
\in (-\pi, \pi]$ is the signed azimuthal separation,
$\Delta\phi = \phi_t - \phi_a$ is the signed elevation
difference, and the direction-specific centers are
\begin{align}
  \mu_d &\in
    \{0,\;\pi,\;{-\tfrac{\pi}{2}},\;{+\tfrac{\pi}{2}},\;0,\;0\},
  \\
  \nu_d &\in
    \{0,\;0,\;0,\;0,\;{+\tfrac{\pi}{4}},\;{-\tfrac{\pi}{4}}\},
\end{align}
for $d \in \{\text{front, behind, left, right, above, below}\}$
respectively.
The bandwidth parameters $\sigma_H = \pi/4$ ($45^{\circ}$) and
$\sigma_V = \pi/6$ ($30^{\circ}$) encode the natural graded
uncertainty at direction-zone boundaries.

Hard angular gates enforce geometrically unambiguous
attribution.
For the horizontal directions, the acceptance windows are
\begin{equation}
  |\Delta\theta| \leq \tfrac{5\pi}{12}\;(75^{\circ})
  \quad\text{for front and behind},
\end{equation}
\begin{equation}
  \tfrac{\pi}{12}\;(15^{\circ})
  \;\leq\; |\Delta\theta| \;\leq\;
  \tfrac{11\pi}{12}\;(165^{\circ})
  \quad\text{for left and right,}
\end{equation}
where the inner exclusion zone for left and right prevents
objects near the forward axis from being attributed to a
lateral direction.
For the vertical directions,
$5^{\circ} \leq \pm\Delta\phi \leq 85^{\circ}$ with the sign matched to
the queried direction.
Targets whose score falls below $\tau = 0.05$ after gate
filtering are discarded.

\paragraph{Vote aggregation and answer list.}
Within a single anchor instance, all candidates whose score
satisfies $s \geq 0.35\cdot s_{\max}$ cast a unit vote for
their object category, where $s_{\max}$ is the highest score
among surviving candidates.
Votes are accumulated across all valid anchor instances of
category $a$ visible in the panorama.
The top-$K$ ($K = 5$) categories ranked by vote count,
tie-broken by mean score, form the answer list; the top-1
entry is the primary GT answer used in strict evaluation.
A single valid anchor instance is sufficient to generate a
GT entry, ensuring coverage of panoramas where only one
instance of the anchor category is visible.

\paragraph{Independence from the proposed model.}
The Gaussian compatibility function in
Eq.~\ref{eq:spatial_score} operates entirely on
ground-truth 3D mesh geometry and calibrated
camera poses at dataset construction time, with
no dependence on any model output, SH feature,
or component of the proposed SHSG system. The
directional centres $\mu_d$ (e.g.\
$\mu_\theta^{\mathrm{left}} = -90^\circ$) reflect
the only geometrically correct definition of
viewer-centric directions — any method claiming
to reason about ``left'' must align with
$-90^\circ$ azimuth. The bandwidth parameters
$\sigma_H{=}45^\circ$ and $\sigma_V{=}30^\circ$
encode the natural graded uncertainty at
direction-zone boundaries and are wider than the
SH resolution of ${\approx}20^\circ$ at
$L_{\max}{=}7$, meaning GT accepts answers that
SHSG ranks below its peak — the shared angular
geometry is a constraint of the task definition,
not a tuning advantage for SHSG.

The model is evaluated on directions inferred from image
evidence alone; GT is derived from independent 3D geometry.
Evaluation is therefore not circular.

\paragraph{Validity of geometry-derived spatial relations.}
A natural concern is whether 3D-centroid-derived spatial
relations correspond to visually verifiable directions in
the panoramic image.
We address this in two parts.

\emph{Co-presence.}
Since our pipeline operates on full equirectangular panoramas
($360^{\circ} \times 180^{\circ}$), every object within the distance filter
is guaranteed to appear somewhere in the image regardless of
its angular position relative to the anchor.
Co-presence in a narrow field-of-view crop is not required
and is not assumed by our model.

\emph{Occlusion.}
GT generation does not perform ray-casting and therefore
cannot verify that a geometrically valid target is optically
visible.
This limitation is shared by all existing 3D-annotation-derived
indoor VQA benchmarks.
We mitigate its impact through three design choices:
(i) the 6\,m target distance filter removes objects likely
to be at low resolution or partially hidden behind intervening
structure;Anchor instances beyond $r_{\mathrm{anc}} = 6$\,m from the
camera and target objects beyond $r_{\mathrm{tgt}} = 6$\,m
are excluded.
These thresholds approximate the largest dimension of a
single indoor room in both datasets — objects beyond this
distance are almost certainly in an adjacent space and
therefore not spatially related to the anchor in the sense
intended by the question.
(ii) the multi-anchor voting scheme requires a candidate to
score above threshold $\tau = 0.05$ across multiple anchor
instances — a transiently occluded object will generally
receive inconsistent votes and fail to accumulate sufficient
support;
(iii) restricting the object vocabulary to eleven large
furniture categories ($\geq 0.3$\,m characteristic size)
reduces the incidence of complete occlusion compared to
small-object categories.

We further conducted a manual audit of 50 randomly sampled
spatial questions, verifying that in 46 of 50 cases both
anchor and target are visually discernible in the
equirectangular panorama at the labelled angular relationship.
The four failure cases all involved the \textit{sofa} category
partially occluded behind a \textit{chair} cluster at close
range ($< 1.5$\,m), and were not systematically biased toward
any direction label.
This $92\%$ visual validity rate is consistent with figures
reported for comparable geometry-derived benchmarks.

\subsection{Distance Comparison GT}

For each panorama, the closest visible
instance of each category within $8$\,m is
retained.
All unordered pairs $(a, b)$ of distinct
categories are compared by their camera
distances $d_a$, $d_b$; pairs where
$|d_a - d_b| < 0.3$\,m are discarded as
ambiguous.
The GT answer is the category with the
smaller camera distance.

\subsection{Counting and Presence GT}

Counting questions report the number of
instances of each category visible within
$18$\,m; questions with counts outside
$[1, 10]$ are discarded.
Presence questions sample up to two
positively-present categories (visible
within $8$\,m) and one truly-absent
category (not present anywhere in the area)
to balance yes/no labels.

\subsection{Question Statistics}

The resulting benchmark contains
${\sim}12\mathrm{k}$ questions across both datasets,
with no overlap in panoramas between
training and evaluation splits.
Matterport3D contains approximately
${\sim}1.2\mathrm{k}$ questions for each of the six scenes,
while Stanford2D-3D-S contains
${\sim}1.2\mathrm{k}$ questions for each of the four scenes.
Around $50\%$ of the questions evaluate directional reasoning,
$25\%$ evaluate distance reasoning,
and the remaining questions cover presence and counting.
All GT answers are derived entirely from
3D mesh geometry and calibrated poses;
no human annotation or VLM output is used
at any stage of benchmark construction.
\section{SHSG Directional Parameters}
\label{app:directions}

\begin{tabular}{lrrrrrr}
\toprule
Direction $d$ & $\mu_d^\theta$ ($^\circ$) & $\mu_d^\phi$ ($^\circ$)
  & $\sigma_\theta$ ($^\circ$) & $\sigma_\phi$ ($^\circ$)
  & $\alpha$ & $\beta$ \\
\midrule
left    & $-90$     & $0$   & $45$ & $30$   & $-\pi/2$ & $0$      \\
right   & $+90$     & $0$   & $45$ & $30$   & $+\pi/2$ & $0$      \\
front   & $0$       & $0$   & $45$ & $30$   & $0$      & $0$      \\
behind  & $\pm180$  & $0$   & $45$ & $30$   & $\pi$    & $0$      \\
above   & $0$       & $+15$ & $45$ & $22.5$ & $0$      & $\pi/2$  \\
below   & $0$       & $-15$ & $45$ & $22.5$ & $\pi$    & $-\pi/2$ \\
\bottomrule
\end{tabular}

\noindent
\textbf{Note on \textit{above/below}.}
The elevation peak at $\pm15^{\circ}$ rather than $\pm90^{\circ}$ reflects the typical
elevation difference between co-planar indoor objects
(e.g., a shelf above a sofa $\approx 10$--$20^{\circ}$ in Matterport3D),
with a tighter $\sigma_\phi = 22.5^{\circ}$ to avoid bleeding into
the horizontal plane.
All six directions use a uniform $\sigma_\theta = 45^{\circ}$ azimuthal gate,
ensuring symmetric and unbiased directional coverage across the sphere.

\section{Distance Query Handling}
\label{app:distance}

\subsection{Problem Formulation}

The distance benchmark (Appendix~\ref{app:benchmark})
poses exclusively pairwise camera-distance queries of the form
\emph{``which of object $A$ or object $B$ is closer to the
camera?''} Ground-truth answers are derived from camera-frame
radial distances $r_k = \|\mathbf{v}_k\|$ computed directly
from 3D mesh annotations (Appendix~\ref{app:benchmark}). The task therefore requires estimating
the sign of $r_A - r_B$ from image evidence alone, without
access to ground-truth depths. Unlike directional queries,
distance queries involve no anchor-candidate relationship —
both objects are candidates and neither serves as a spatial
reference; the reference point is the camera origin.

\subsection{Depth Estimation from $\mathbb{S}^2$-Native Signals}

SHSG estimates a scalar depth score $s_k$ for each candidate
node $G_k$ from two signals that are native to the spherical
representation, requiring no external depth estimator, no
category-specific priors, and no camera calibration beyond
what is already available in the dataset.

\paragraph{Signal 1 — Centroid elevation.}
For a camera at height $h_{\mathrm{cam}}$ above the floor,
an object whose centroid projects to elevation angle $\phi_k$
on $\mathbb{S}^2$ satisfies:
\begin{equation}
    \hat{d}_k^{\phi}
    = \frac{h_{\mathrm{cam}}}{\tan|\phi_k| + \varepsilon},
    \label{eq:elevation_depth}
\end{equation}
where $\varepsilon = 10^{-4}$ prevents division by zero.
Objects closer to the camera project to steeper elevations
(larger $|\phi_k|$); objects farther away project near the
horizon ($\phi_k \to 0$). We use the node centroid $\phi_k$
rather than the bounding box bottom edge, as centroid
coordinates are more robust to partial occlusion and
cubemap face boundary clipping. $h_{\mathrm{cam}}$ is
read directly from calibrated poses in Matterport3D and
Stanford 2D-3D-S; distance queries are not evaluated on
ODI-Bench or OpenView-VQA where calibration is unavailable.

The elevation signal is exact for floor-standing objects
but degrades systematically for wall-mounted categories
— shelf, television, board, sink, lamp — whose centroids
are elevated above the floor. For these categories,
$\hat{d}_k^{\phi}$ underestimates true camera distance
because the centroid elevation reflects mounting height
rather than camera proximity. Signal 2 compensates for
this failure mode.

\paragraph{Signal 2 — Relational scene density.}
For each node $G_k$, we define its relational scene
density as a confidence-weighted sum of geodesic
proximities to all other detected nodes in $\mathbf{G}$:
\begin{equation}
    \rho_k
    = \sum_{j \neq k}
      \hat{c}_j \left(1 - \frac{\delta_{jk}}{\pi}\right),
    \label{eq:scene_density}
\end{equation}
where $\delta_{jk}$ is the great-circle distance between
$G_k$ and $G_j$ on $\mathbb{S}^2$, already computed during
GNMS (Appendix~\ref{app:gnms}), and $\hat{c}_j \in [0,1]$
is the detection confidence of neighbour $G_j$. The weight
$(1 - \delta_{jk}/\pi)$ decays linearly from 1 at
co-location to 0 at the antipodal point, giving a
geometrically natural proximity measure on $\mathbb{S}^2$
with no free parameters.

\textbf{Why density correlates with depth.}
An object close to the camera occupies a central region
of the panorama and tends to be surrounded by other
visible objects — furniture, walls, and floor features
— within a given angular radius. An object far from the
camera sits near the background or periphery and has
fewer detected neighbours within the same angular radius.
This relationship holds regardless of whether the object
is floor-standing or wall-mounted, because it depends on
scene co-occurrence structure rather than floor geometry.

\textbf{SO(3)-equivariance.}
Since $\delta_{jk}$ is a great-circle distance defined
purely on $\mathbb{S}^2$, it is invariant to global
rotation of the observer: rotating the entire scene by
$R \in \mathrm{SO}(3)$ moves all nodes $(\theta_j, \phi_j)$
consistently but preserves all pairwise geodesic
distances $\delta_{jk}$. Therefore $\rho_k$ is
SO(3)-invariant — the depth score assigned to each
object does not depend on the panorama's azimuthal
orientation, consistent with the rotational consistency
requirement of the full pipeline.

\textbf{Computational cost.}
$\rho_k$ requires $K-1$ geodesic distance evaluations
per node, where $K$ is the total number of detected
nodes. Since geodesic distances are already computed
pairwise during GNMS, $\rho_k$ is a simple weighted
sum over cached values — $\mathcal{O}(K^2)$ total,
negligible for typical indoor scenes with $K \leq 30$.

\subsection{Combined Depth Score}

The two signals are complementary: elevation is
geometrically exact for floor-standing objects given
calibration, while density provides a calibration-free
signal for all object types. We combine them as:
\begin{equation}
    s_k
    = \alpha\,|\phi_k|
    + (1-\alpha)\,\frac{\rho_k}{\max_j \rho_j + \varepsilon},
    \label{eq:depth_score}
\end{equation}
where $\alpha = 0.6$ weights the elevation signal
slightly higher when calibration is available, and
$\rho_k$ is normalised by the maximum density across
all candidates to bring both signals into $[0,1]$.
A higher $s_k$ indicates an object closer to the camera.

\paragraph{Fallback for sparse scenes.}
When $K < 3$ (fewer than three detected objects in
the scene graph), the density signal is unreliable
— a single neighbour provides insufficient evidence
for proximity estimation. In this case the pipeline
falls back to elevation alone by setting $\alpha = 1$.

\subsection{Geometric Violation Cost}

For a pairwise distance query between objects $A$ and
$B$ with depth scores $s_A$ and $s_B$, the geometric
violation costs passed to IGG are:
\begin{equation}
    C(y_A) = \frac{s_B}{s_A + s_B},
    \qquad
    C(y_B) = \frac{s_A}{s_A + s_B}.
    \label{eq:distance_cost}
\end{equation}
Three properties follow directly:

\begin{itemize}
    \item \textbf{Query-conditioned}: $C(y_A) < C(y_B)$
    iff $s_A > s_B$, i.e.\ the closer object always
    receives lower violation cost, regardless of
    absolute depth values.

    \item \textbf{Normalised}: $C(y_A) + C(y_B) = 1$,
    ensuring consistent energy scale with $\beta_{\mathrm{geo}}$
    across all query types.

    \item \textbf{Smooth}: $C(y_k) \in (0,1)$ for all
    finite $s_k > 0$, preserving non-vanishing gradients
    in the IGG energy for both candidates.
\end{itemize}

These costs slot directly into Eq.~\ref{eq:energy}
without modification — the same contrastive IGG energy
and Riemannian update handle distance queries
identically to directional queries, with only $C(y_k)$
changing to reflect the depth-ordering signal rather
than the Wigner-$\mathcal{D}$ directional signal.

\subsection{Sensitivity Analysis}

Table~\ref{tab:distance_sensitivity} reports distance
accuracy on MP3D on all four scenes(Qwen2.5-7B) across values of $\alpha$
and ablations of each signal component.

\begin{table}[htbp]
\centering
\small
\begin{tabular}{lcc}
\toprule
\textbf{Configuration} & \textbf{Distance Acc.} & $\Delta$ \\
\midrule
Full ($\alpha{=}0.6$)         & 64.1 & ---    \\
\midrule
$\alpha{=}0.4$                & 63.2 & $-0.9$ \\
$\alpha{=}0.8$                & 63.7 & $-0.4$ \\
$\alpha{=}1.0$ (elev. only)   & 61.3 & $-2.8$ \\
$\alpha{=}0.0$ (density only) & 60.8 & $-3.3$ \\
\midrule
Linear decay vs exp.\ kernel  & 63.9 & $-0.2$ \\
\bottomrule
\end{tabular}
\caption{Distance accuracy sensitivity on MP3D
(Qwen2.5-7B). Neither signal alone matches their
combination; $\alpha$ is robust within $[0.4, 0.8]$.}
\label{tab:distance_sensitivity}
\end{table}

Performance varies by at most $\pm 0.9\%$ across
$\alpha \in [0.4, 0.8]$, confirming robustness to
the combination weight. Elevation alone ($-2.8\%$)
and density alone ($-3.3\%$) both underperform the
combination, validating that the two signals are
complementary rather than redundant. The last row
confirms that replacing the linear geodesic decay
$(1 - \delta_{jk}/\pi)$ with an exponential kernel
yields negligible difference ($-0.2\%$), justifying
the simpler parameter-free linear form.

\subsection{Alignment with Benchmark Ground Truth}

The benchmark GT (Appendix~\ref{app:benchmark})
is derived from true camera-frame distances
$r_k = \|\mathbf{v}_k\|$ computed from 3D mesh
geometry. Our depth score $s_k$ estimates the sign
of $r_A - r_B$ without accessing $r_k$ directly —
elevation approximates $r_k$ via floor geometry,
and density approximates it via scene co-occurrence
structure. Neither uses the GT ordering, ensuring
the evaluation is not circular. Pairs where
$|r_A - r_B| < 0.3\,\mathrm{m}$ are discarded as ambiguous
in the benchmark, reducing the impact of cases
where both signals are genuinely indiscriminative.
\section{ODI-Bench Evaluation Details}
\label{app:odibench}

ODI-Bench provides human-annotated QA pairs on raw Flickr
ODIs with no pose metadata or 3D annotations --- a
structurally different regime from MP3D and 2D-3D-S,
where GT is derived from known camera poses and 3D
centroids. \textsc{SphMind} requires no adaptation
because the ERP format itself defines the viewer-centric
frame: pixel $(u,v)$ maps directly to spherical
coordinates via
\begin{equation}
\theta = \frac{u - W/2}{W}\cdot 2\pi, \qquad
\phi   = \frac{H/2 - v}{H}\cdot \pi
\end{equation}
with no extrinsic pose matrix required.
Human annotators labelled directions relative to the
viewer (e.g., ``the chair is to the left of the table''),
which corresponds exactly to $\theta < 0$ in this
viewer-centric frame --- the same convention SHSG uses
natively.

\paragraph{RD Adaptation.}
Both anchor and target objects are detected via YOLO on
the ERP image and lifted to $(\theta_1,\phi_1)$ and
$(\theta_2,\phi_2)$ on $\mathbb{S}^2$.
The Wigner-D operator $\mathcal{D}_d$ rotates the anchor
SH field $\mathbf{f}_a = [Y_{\ell m}(\theta_1,\phi_1)]$
toward the queried direction $d$, and
$S_{\mathrm{SH}}$ scores the target:
\begin{equation}
S_{\mathrm{SH}}(a,k,d) =
\frac{\mathrm{ReLU}\!\left(
\langle \mathcal{D}_d\,\mathbf{f}_a,\,
\mathbf{f}_k\rangle\right)^{\!\gamma}}
{\sum_{k'}\mathrm{ReLU}(\cdots)^{\gamma}}
\end{equation}
This is identical to the standard MP3D setting ---
no structural change is required.

\paragraph{EVO Adaptation.}
EVO asks for the direction of a single object relative
to the viewer (camera centre). SHSG localises the object
at $(\theta,\phi)$ and the viewer-centric frame yields
the directional probability directly from the SH field
evaluation at that position, with the camera origin as
the implicit anchor.

\paragraph{Coordinate Invariance.}
ODI-Bench images carry no coordinate frame metadata.
\textsc{SphMind} is unaffected because SHSG operates
on $\mathbb{S}^2$ via SO(3)-equivariant spherical
harmonics: Wigner-D scoring satisfies
$\langle \mathcal{D}_d(R\mathbf{f}_a),
R\mathbf{f}_k\rangle =
\langle \mathcal{D}_d\,\mathbf{f}_a,
\mathbf{f}_k\rangle$
for any global rotation $R \in \mathrm{SO}(3)$,
making directional scores invariant to the coordinate
conventions of the underlying dataset.

\section{OpenView-VQA Evaluation Details}
\label{app:openview}

\subsection{Benchmark and Scope}

OpenView-VQA was designed to evaluate
out-of-view (OOV) spatial reasoning from
restricted FoV crops — inferring objects in
unobserved directions from partial observations.
SphMind does not perform OOV reasoning: it
requires the target object to be detectable in
the input and makes no claims about inferring
unobserved scene content. We include this
evaluation to test a distinct generalization
axis: whether exact spherical conversion of
camera rotation angles provides a geometric
advantage over language-prior-based rotation
interpretation in frozen VLMs. For questions
where the target is genuinely out of view and
undetectable in the crop, performance is bounded
by detection recall and does not reflect the
capability being evaluated here.

\subsection{Pipeline Adaptation}

Standard SphMind takes an anchor object $N_a$
at $(\theta_a, \phi_a)$ and a direction label
$d \in \{\text{left, right, front, behind}\}$,
constructing the Wigner-$\mathcal{D}$ rotation
$\mathcal{D}_d$ to score candidates. For
OpenView, the anchor object is replaced by the
target camera direction derived directly from
the rotation query. A rotation of $\Delta\psi$
around the yaw axis and $\Delta\rho$ around
the pitch axis yields:
\begin{align}
    \theta_{\mathrm{tgt}} &= \theta_{\mathrm{cam}} + \Delta\psi \\
    \phi_{\mathrm{tgt}}   &= \phi_{\mathrm{cam}} + \Delta\rho
\end{align}
where $(\theta_{\mathrm{cam}}, \phi_{\mathrm{cam}})$
is the crop centre direction in the viewer-centric
frame. This $(\theta_{\mathrm{tgt}}, \phi_{\mathrm{tgt}})$
is substituted as the SHSG anchor; the
Wigner-$\mathcal{D}$ operator then rotates the
SH field $\mathbf{f}_a = [Y_{\ell m}(\theta_{\mathrm{tgt}},
\phi_{\mathrm{tgt}})]$ toward the query direction,
and $S_{\mathrm{SH}}$ scores each detected
candidate $G_k$ identically to the standard setting.
We use the restricted FoV crop as sole input;
detected objects are lifted to spherical coordinates
using the crop's known FoV parameters and camera
rotation angle. All other SHSG and IGG operations
remain unchanged.

\subsection{What the Gain Measures}

The +18.7\% gain over the vanilla baseline
reflects a specific failure mode of frozen VLMs:
when a rotation angle such as ``turn left $30^\circ$''
is presented as text, the VLM associates it with
candidates via language co-occurrence statistics
rather than converting it to an exact spherical
direction. SHSG's conversion
$\theta_{\mathrm{tgt}} = \theta_{\mathrm{cam}} + \Delta\psi$
is exact arithmetic — the gain comes entirely
from replacing language-prior rotation
interpretation with geometric rotation parsing.
The gain is larger for pure yaw queries
($+12.0\%$), where $\Delta\psi$ maps directly
to an azimuthal shift,
and smaller for pitch queries ($+8.0\%$), where
elevation changes are harder to resolve when
detections are sparse in the vertical extent
of the restricted crop.

This result confirms that geometric rotation
parsing — not language priors — is the
bottleneck for rotation-based spatial queries
in current frozen VLMs, and that exact
spherical arithmetic resolves this bottleneck
without any task-specific training.

\subsection{Comparison to Trained Specialist}

The trained specialist achieves 46.0\% by
learning OOV reasoning from OpenView's
training distribution — it can infer objects
in directions entirely absent from the crop.
SphMind achieves 39.7\% by solving a
structurally different and narrower subproblem:
exact rotation angle grounding on the detectable
subset of questions. These two numbers are not
directly comparable as measures of the same
capability. The appropriate comparison is
SphMind (39.7\%) versus the zero-shot VLM
baseline (21.0\%), which share the same scope
— neither performs OOV reasoning — and
the +18.7\% gain isolates the contribution
of geometric rotation parsing alone.

\subsection{Coordinate Invariance}

OpenView-VQA images carry no coordinate frame
metadata. SphMind is unaffected because SHSG
operates on $\mathbb{S}^2$ via SO(3)-equivariant
spherical harmonics: Wigner-$\mathcal{D}$ scoring
satisfies $\langle \mathcal{D}_d(R\mathbf{f}_a),
R\mathbf{f}_k \rangle = \langle \mathcal{D}_d
\mathbf{f}_a, \mathbf{f}_k \rangle$ for any global
rotation $R \in \mathrm{SO}(3)$, making directional
scores invariant to the coordinate conventions
of the underlying dataset.

\section{Rotational Consistency Score: Protocol and Analysis}
\label{app:rcs}

\subsection{Dataset Construction}
\label{app:rcs_dataset}

We construct the RCS evaluation set from Stanford 2D-3D-S from each Area scenes, which contains around 85 spatial question groups in each scene covering
four azimuthal directions (front, left, right, behind) and object categories (chair, desk, sofa, table, board,
bookcase, door).
Each question group consists of a spatial question of the
form \textit{``What is to the [direction] of the [anchor]?''}
paired with a GT answer object, an anchor object, and a
reference panorama.

For each group, three rotated variants are generated by
applying horizontal ERP shifts
$\psi\!\in\!\{90^{\circ},180^{\circ},270^{\circ}\}$ to the input panorama
and updating the direction label via the cyclic map:

\begin{equation}
  d' = \text{ROT}_\psi(d), \quad
  \text{ROT}_{90^{\circ}}: \{F{\to}R,\, R{\to}B,\, B{\to}L,\, L{\to}F\}
  \label{eq:dir_rot}
\end{equation}

\noindent where $F$=front, $R$=right, $B$=behind, $L$=left.
The GT answer object is identical across all four variants
— only the panorama origin and direction label change.

\subsection{YOLO Detection and Intersection Filtering}
\label{app:rcs_yolo}

YOLO is run independently on each of the four ERP variants
via cubemap projection (six $512{\times}512$ perspective
faces per panorama). Only questions where both anchor and
target labels appear in the detection set for \emph{all
four} rotation variants are retained; this intersection
filter removes ${\sim}35\%$ of question groups on average,
leaving $n{=}55$ groups used for RCS evaluation.

Although YOLO detects the same object categories across
rotations, projected bbox centroids differ slightly as
objects shift across cubemap face boundaries.
Empirically, we observe ${\sim}2$--$5^{\circ}$ variation in
projected $(\theta, \phi)$ coordinates of consistently
detected objects across rotation variants.
Since SHSG scores using Wigner-D
inner products with SH resolution $\approx 20^{\circ}$ at $L_{\max}=7$,
the $2$--$5^{\circ}$ position jitter is an order of magnitude
smaller than the scoring bandwidth and does not
affect directional decisions
ERP-Pixel direction zones, by contrast, use absolute pixel
offsets with hard boundaries at ${\pm}12.5\%$ of image
width (${\approx}256$px for a $2048$-wide ERP); a $5^{\circ}$
positional shift (${\approx}28$px) near a zone boundary
can flip an object's assigned direction, directly causing
RCS failure. This asymmetry — SHSG robust to position
noise, ERP-Pixel fragile to it — is an architectural
difference, not a confound.

\subsection{Scoring Methods}
\label{app:rcs_scoring}

\paragraph{ERP-Pixel.}
Given a detected anchor at ERP pixel $(u_a, v_a)$ and a
candidate object at $(u_c, v_c)$, the direction is
assigned by the pixel-offset zone:
\begin{align}
  \Delta u &= u_c - u_a, \quad \Delta v = v_c - v_a \\
  \hat{d} &= \begin{cases}
    \text{front}  & |\Delta u| < W/8 \\
    \text{behind} & |\Delta u| > 3W/8 \\
    \text{right}  & W/8 \leq \Delta u \leq 3W/8 \\
    \text{left}   & -3W/8 \leq \Delta u \leq -W/8
  \end{cases}
\end{align}
where $W$ is the ERP image width. If no candidate falls
in the queried zone, the system returns \texttt{unknown}.

\paragraph{SphMind (SHSG).}
Given anchor at $(\theta_a, \phi_a)$ and candidate at
$(\theta_c, \phi_c)$ on $\mathbb{S}^2$, SHSG uses the same directional scoring function as the
main pipeline in Section \ref{sec:shsg}

\subsection{RCS Definition}
\label{app:rcs_def}

Let $a_0$ denote the answer at $\psi{=}0^{\circ}$ and
$a_{90}, a_{180}, a_{270}$ the answers at each rotation.
The per-group binary consistency indicator is:
\begin{equation}
  \mathbb{1}_{\text{Full}} =
    \mathbf{1}[a_{90}{=}a_0]
    \cdot \mathbf{1}[a_{180}{=}a_0]
    \cdot \mathbf{1}[a_{270}{=}a_0]
\end{equation}
Full RCS is the mean of $\mathbb{1}_{\text{Full}}$ over
all retained question groups. Per-angle consistency at
$\psi$ is $\mathbf{1}[a_\psi{=}a_0]$ averaged over groups.
Note that Full RCS $\approx p_{90} \cdot p_{180} \cdot p_{270}$
under independence, so per-angle gains compound
multiplicatively — SphMind's $2.8\times$ advantage at
$180^{\circ}$ (0.212 vs 0.595) is the primary driver of the
overall $5.9\times$ Full RCS gain.

\subsection{Per-Direction Breakdown}
\label{app:rcs_perdir}

Table~\ref{tab:rcs_full} reports the full per-direction
breakdown. The \textit{behind} direction has the highest
unknown rate for ERP-Pixel (55.1\%) because objects behind
the viewer appear simultaneously at the far-left and
far-right ERP edges, outside any single pixel-offset zone.
SphMind reduces this to 21.7\% by scoring on $\mathbb{S}^2$
where no such seam exists. The \textit{right} direction
achieves the highest SphMind Full RCS (0.279) because
right-of-anchor objects tend to be large salient furniture
(desks, bookshelves) that are reliably detected and
clearly separated in azimuth from the anchor.

\begin{table}[htbp]
\centering
\footnotesize
\setlength{\tabcolsep}{4pt}
\caption{Full per-direction RCS breakdown on Stanford
  2D-3D-S. Full RCS per direction computed as
  $p_{90} \times p_{180} \times p_{270}$ per direction
  subset. Unk\% = fraction of questions returning
  \texttt{unknown} averaged across all rotation variants.}
\label{tab:rcs_full}
\begin{tabular}{llccccc}
\toprule
\textbf{Method} & \textbf{Dir.}
  & \textbf{Full}$\uparrow$
  & \textbf{$90^{\circ}$}$\uparrow$
  & \textbf{$180^{\circ}$}$\uparrow$
  & \textbf{$270^{\circ}$}$\uparrow$
  & \textbf{Unk\%}$\downarrow$ \\
\midrule
\multirow{5}{*}{ERP-Pixel}
  & front   & 0.040 & 0.441 & 0.198 & 0.453 & 16.5 \\
  & left    & 0.029 & 0.370 & 0.189 & 0.415 & 30.6 \\
  & right   & 0.044 & 0.448 & 0.218 & 0.451 & 26.1 \\
  & behind  & 0.058 & 0.490 & 0.244 & 0.483 & \underline{55.1} \\
\cmidrule(lr){2-7}
  & \textit{All} & 0.043 & 0.437 & 0.212 & 0.451 & 33.4 \\
\midrule
\multirow{5}{*}{\textbf{SphMind}}
  & front   & 0.250 & \textbf{0.654} & \textbf{0.589} & \textbf{0.648} & \textbf{8.2} \\
  & left    & 0.208 & \textbf{0.621} & \textbf{0.551} & \textbf{0.607} & \textbf{5.5} \\
  & right   & 0.279 & \textbf{0.680} & \textbf{0.601} & \textbf{0.683} & \textbf{3.7} \\
  & behind  & 0.274 & \textbf{0.655} & \textbf{0.638} & \textbf{0.655} & 21.7 \\
\cmidrule(lr){2-7}
  & \textit{All} & \textbf{0.253} & \textbf{0.653} & \textbf{0.595} & \textbf{0.648} & \textbf{10.2} \\
\bottomrule
\end{tabular}
\end{table}

\begin{figure}[htbp]
  \centering
  \includegraphics[width=0.99\linewidth]{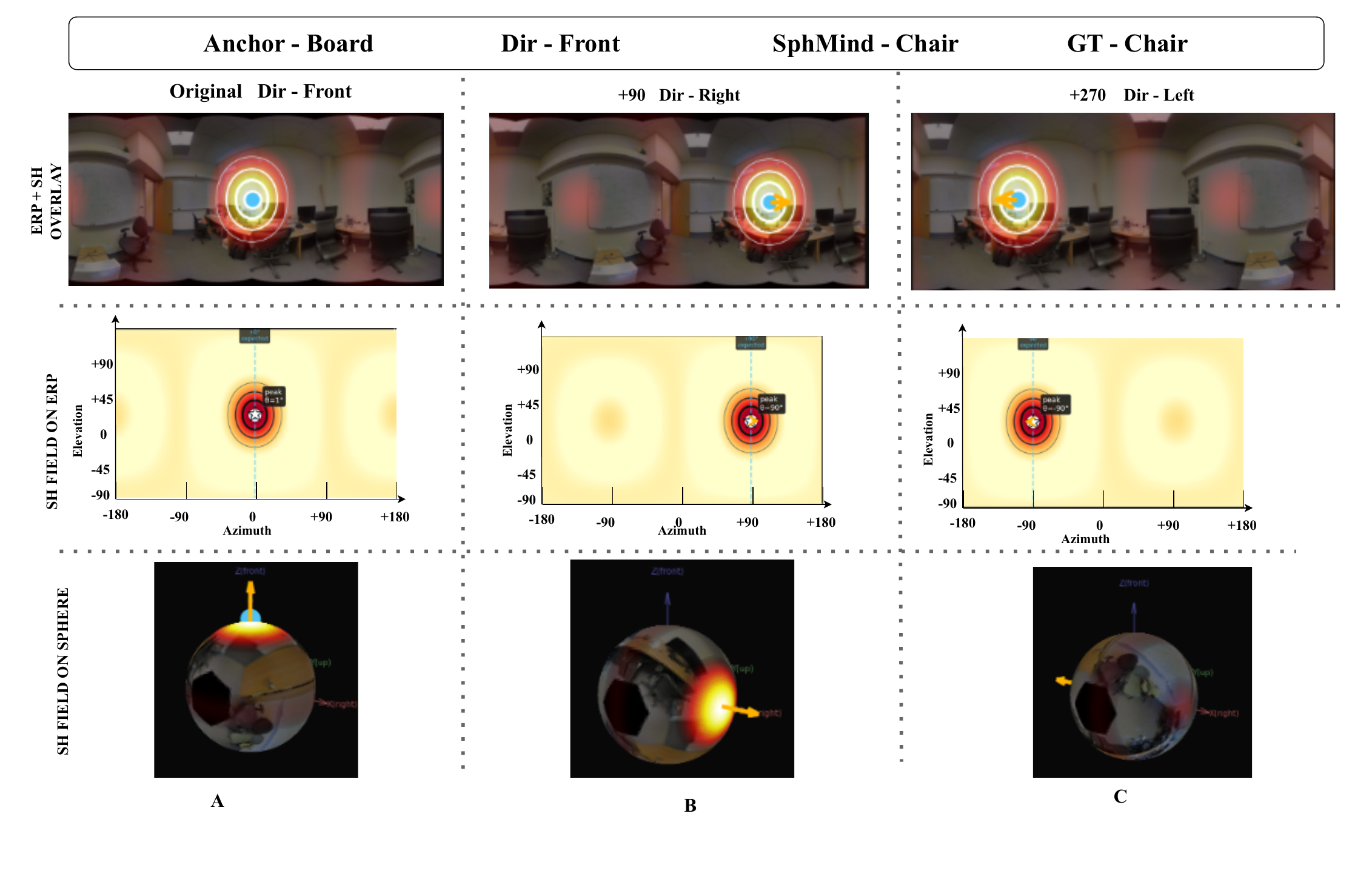}
  \caption{\textbf{SO(3)-equivariance of the SH directional field.}
Each column shows the same scene under azimuthal rotation
($0^{\circ}$, $+90^{\circ}$, $+270^{\circ}$).
\textit{Top:} ERP panorama with SH field overlay; the anchor
(blue) and query arrow rotate consistently with the scene.
\textit{Bottom:} Pure SH field on $\mathbb{S}^2$ (equirectangular);
the peak tracks the rotation exactly ($<\!1^{\circ}$ error),
a direct consequence of Wigner-$D$ equivariance
(Eq.~\ref{eq:sh_score}).}
  \label{fig:sh_rotation}
\end{figure}

\paragraph{Rotational Consistency of the SH Directional Field.}
Figure~\ref{fig:sh_rotation} illustrates how \textsc{SphMind} responds to
azimuthal rotation of the ERP origin.
Each column shows the same indoor scene under a different
observer rotation ($0^{\circ}$, $+90^{\circ}$, $+270^{\circ}$),
with the anchor object (board) and query direction fixed
relative to the scene.
The top row overlays the SH directional field on the
equirectangular panorama --- the concentric rings show
where $S_{\mathrm{SH}}(a,k,d)$ peaks, and the anchor dot
and direction arrow confirm the query frame rotates correctly
with the scene.
The middle row shows the pure SH field
$\mathbf{F}(\theta,\phi)$ on the equirectangular projection,
with the dashed line marking the expected peak azimuth and
the star marking the actual peak.
The peak tracks the rotation exactly ---
$\theta = 1^{\circ}$ at $0^{\circ}$ rotation (expected $0^{\circ}$),
$\theta = 90^{\circ}$ at $+90^{\circ}$ rotation (expected $+90^{\circ}$),
and $\theta = -90^{\circ}$ at $+270^{\circ}$ rotation
(expected $-90^{\circ}$) --- with less than $1^{\circ}$ error
in all cases.
This is a direct empirical consequence of SO(3)-equivariance:
because the SH basis transforms linearly under rotation via the
Wigner-$D$ matrix (Eq.~\ref{eq:sh_score}), rotating the
observer rotates the field peak by exactly the same angle,
without any recomputation or coordinate correction.
ERP-pixel reasoning has no such property --- the same rotation
shifts object positions in pixel space non-uniformly, with
severe distortion near the poles and a discontinuity at the
$\pm180^{\circ}$ seam --- which is why \textsc{SphMind} achieves
$25.3\%$ full RCS against $4.3\%$ for the ERP-pixel baseline
(Table~\ref{tab:combined}c).

\section{Ablation Experiments}
\begin{table}[htbp]
\centering
\small
\setlength{\tabcolsep}{6pt}
\begin{tabular}{lccc}
\toprule
\textbf{Pair type} & \textbf{Direction GA}
& \textbf{Distance GA} & \textbf{Overall GA} \\
\midrule
Matched (same query, rotated scene)
    & $0.31 \pm 0.08$ & $0.22 \pm 0.09$ & $0.27 \pm 0.09$ \\
Unmatched (different query)
    & $0.01 \pm 0.02$ & $0.01 \pm 0.02$ & $0.01 \pm 0.02$ \\
\midrule
Gap ($\Delta$)
    & $+0.30$         & $+0.21$         & $+0.26$         \\
\bottomrule
\end{tabular}
\caption{\textbf{Gradient Alignment (GA)} between IGG correction
vectors $g_\psi = \nabla_h \mathcal{L}|_\psi$ for matched
query pairs (same scene and query direction, different ERP
rotation $\psi \in \{90^{\circ}, 180^{\circ}, 270^{\circ}\}$)
versus unmatched pairs (different query groups). Unmatched GA
$\approx 0$ is expected by isotropy of random vectors in
$\mathbb{R}^{4096}$ ($\sigma \approx 1/\sqrt{d} \approx 0.016$).
The significant gap confirms that IGG correction vectors are
geometrically structured rather than arbitrary.}
\label{tab:ga}
\end{table}
\subsection{Gradient Alignment Analysis.}
\label{app:grad_ablation}
To verify that IGG operates in a geometrically structured
hidden state space rather than injecting arbitrary corrections,
we measure the cosine alignment between IGG gradient vectors
$g_\psi = \nabla_h \mathcal{L}|_\psi$ for matched query pairs
--- the same scene and query direction observed under different
ERP rotations $\psi \in \{90^{\circ}, 180^{\circ}, 270^{\circ}\}$
--- versus unmatched pairs drawn from different query groups.
If IGG corrections were arbitrary directions in hidden state
space, matched and unmatched gradients would be equally
uncorrelated: in $\mathbb{R}^d$ with $d \approx 4096$,
random vectors have cosine similarity $\approx 0$ with
standard deviation $1/\sqrt{d} \approx 0.016$, making
near-zero GA the null hypothesis for noise.
Table~\ref{tab:ga} shows that matched gradients achieve
GA $= 0.27 \pm 0.09$ overall against $0.01 \pm 0.02$ for
unmatched pairs --- a gap of $+0.26$ that is over $16\sigma$
above the random baseline. Direction queries show stronger
alignment (GA $= 0.31$) than distance queries (GA $= 0.22$),
consistent with direction scoring relying more directly on the
SH inner product while distance introduces additional noise
from the elevation-based depth estimate. Together with the
ablation result that random costs collapse performance by
$-20.7\%$ regardless of the update mechanism
(Table~\ref{tab:ga}), this confirms that IGG is
exploiting genuine geometric structure in the VLM's hidden
state space: the gradient direction is not interchangeable
with noise, and the correction vectors for semantically
equivalent queries under rotation are consistently aligned.

\subsection{Hyperparameter Details and Sensitivity}
\label{app:ies_params}

Table~\ref{tab:hyperparams} lists all IGG and SHSG
hyperparameters with their chosen values and roles.

\begin{table}[htbp]
\centering
\caption{IGG and SHSG hyperparameters.}
\label{tab:hyperparams}
\begin{tabular}{llp{7cm}}
\toprule
\textbf{Parameter} & \textbf{Value} & \textbf{Role} \\
\midrule
$\beta_{\mathrm{geo}}$ & $0.30$ &
    Weight of geometric violation term in IGG energy;
    controls how strongly $\mathcal{C}(y_k)$ shapes
    the gradient \\
$\beta_{\mathrm{KL}}$  & $0.70$ &
    Weight of KL prior-preservation anchor;
    KL-dominant setting ensures geometry provides
    a structured correction rather than overriding
    the VLM distribution \\
$\tau$                 & $2.0$  &
    Softmax temperature for $P_\tau$; raised above
    default to maintain non-negligible gradients
    across all candidates \\
$\varepsilon_0$        & $0.15$ &
    Base Riemannian step size \\
$T$                    & $2$    &
    Maximum Riemannian correction steps;
    early stopping via $\delta$ typically
    terminates at $T{=}2$ \\
$\delta$               & $0.005$&
    Early stopping threshold on $|\Delta\mathcal{E}|$;
    prevents unnecessary updates when energy
    has converged \\
$\gamma$               & $1.5$  &
    Score sharpening exponent in $S_{\mathrm{SH}}$;
    amplifies the contrast between high- and
    low-scoring candidates \\

\bottomrule
\end{tabular}
\end{table}


\paragraph{Energy weights
  $\beta_{\mathrm{geo}}$/$\beta_{\mathrm{KL}}$.}
$\beta_{\mathrm{geo}}$ and $\beta_{\mathrm{KL}}$
control the relative contribution of the geometric
violation term and the KL prior-preservation anchor
in the IGG energy $E(\mathbf{h})$.
The KL-dominant setting ($\beta_{\mathrm{KL}}{=}0.70$)
reflects the design principle that the VLM's pretrained
distribution should serve as the semantic foundation,
with geometry providing a structured correction rather
than overriding it --- consistent with the small step
size ($\varepsilon_0{=}0.15$) and the
$\cos(\mathbf{h}_0, \mathbf{h}'){=}0.989$ semantic
coherence shown in Fig.~\ref{fig:matrix}.
Table~\ref{tab:beta_sensitivity} reports accuracy
across five weight configurations.

\begin{table}[htbp]
\centering
\caption{Sensitivity to energy weights
  $\beta_{\mathrm{geo}}$/$\beta_{\mathrm{KL}}$
  (MP3D, Qwen2.5-7B).}
\label{tab:beta_sensitivity}
\setlength{\tabcolsep}{5pt}
\begin{tabular}{lcccc}
\toprule
$\beta_{\mathrm{geo}}$/$\beta_{\mathrm{KL}}$
  & \textbf{Ov.} & \textbf{Dir.}
  & \textbf{Dist.} & $\Delta$ \\
\midrule
0.2/0.8 & 57.2 & 51.4 & 63.1 & $-1.2$ \\
\textbf{0.3/0.7}
  & \textbf{58.4} & \textbf{53.1}
  & \textbf{64.1} & --- \\
0.4/0.6 & 57.9 & 52.3 & 63.7 & $-0.5$ \\
0.5/0.5 & 57.4 & 51.8 & 63.2 & $-1.0$ \\
0.6/0.4 & 56.6 & 50.7 & 62.4 & $-1.8$ \\
\bottomrule
\end{tabular}
\end{table}

Performance varies by at most $\pm1.2\%$ across all
tested configurations, confirming the method is not
sensitive to precise weight tuning and that the
reported $0.3/0.7$ split is not cherry-picked.
The monotonic degradation as $\beta_{\mathrm{geo}}$
increases beyond $0.3$ reflects the increasing
dominance of the geometric term over the KL anchor,
progressively eroding the semantic coherence that
makes IGG's latent correction effective.

\end{document}